\documentclass[letterpaper]{article}

\usepackage[final]{corl_2022} 

\usepackage{amssymb,amsmath,amsthm}
\usepackage{wrapfig,booktabs}
\usepackage{graphicx}
\usepackage{adjustbox}
\usepackage{booktabs}
\usepackage{multicol}
\usepackage{multirow}
\usepackage{verbatim}
\usepackage{pifont}
\usepackage{xspace}
\usepackage{etoc}
\usepackage{siunitx} 
\usepackage{enumitem} 
\setlist[itemize]{leftmargin=*} 
\usepackage[font={footnotesize}]{caption}

\usepackage{color, soul} 
\renewcommand\hl[1]{#1} 

\renewcommand{\contentsname}{Supplement Contents}
\etocsettocstyle{\section*{\contentsname}}{}%
\newcommand{\beginsupplement}{%
    \setcounter{table}{0}
    \renewcommand{\thetable}{S\arabic{table}}%
    \setcounter{figure}{0}
    \renewcommand{\thefigure}{S\arabic{figure}}%
    \setcounter{section}{0}
 }

\newtheorem{remark2}{Remark}

\newcommand{\remark}[3]{{\color{#2}[#1: #3]}}

\usepackage[usenames,dvipsnames]{xcolor} 

\newcommand{\daniel}[1]{\remark{Daniel}{blue}{#1}}

\newcommand{\toolflow}{ToolFlowNet\xspace}
\newcommand{\pouring}{PourWater\xspace}
\newcommand{\scooping}{ScoopBall\xspace}

\newcommand{\pointloss}{L_{\textrm{point}}}
\newcommand{\consistency}{L_{\textrm{consistency}}}

\newcommand{\cmark}{\textcolor[HTML]{59a14f}{\ding{51}}}%
\newcommand{\xmark}{\textcolor[HTML]{e15759}{\ding{55}}}%

\newcommand{\pcl}{\mathbf{P}} 
\newcommand{\flow}{\mathbf{F}} 
\newcommand{\ba}{\mathbf{a}}

\newcommand{\bo}{\mathbf{o}}

\newcommand{\bR}{\mathbf{R}}
\newcommand{\bt}{\mathbf{t}}
\newcommand{\bT}{\mathbf{T}}

\newcommand{\ie}{i.e.,\xspace}
\newcommand{\eg}{e.g.,\xspace} 

\title{\toolflow: Robotic Manipulation with Tools via Predicting Tool Flow from Point Clouds}

\author{
  Daniel Seita,
  Yufei Wang$^\dagger$,
  Sarthak J. Shetty$^\dagger$,
  Edward Yao Li$^\dagger$,
  Zackory Erickson,
  David Held\\
  $^\dagger$Equal contribution.\\
  The Robotics Institute, Carnegie Mellon University, USA\\
  Correspondence to: \texttt{dseita@andrew.cmu.edu}
}

\begin{document}
\maketitle


\begin{abstract}
Point clouds are a widely available and canonical data modality which convey the 3D geometry of a scene. Despite significant progress in classification and segmentation from point clouds, policy learning from such a modality remains challenging, and most prior works in imitation learning focus on learning policies from images or state information.
In this paper, we propose a novel framework for learning policies from point clouds for robotic manipulation with tools. We use a novel neural network, \toolflow, which predicts dense per-point flow on the tool that the robot controls, and then uses the flow to derive the transformation that the robot should execute. 
We apply this framework to imitation learning of challenging deformable object manipulation tasks with continuous movement of tools, including scooping and pouring, and demonstrate significantly improved performance over baselines which do not use flow. We perform \hl{50} physical scooping experiments with \toolflow and attain \hl{82}\% scooping success.
See \url{https://tinyurl.com/toolflownet} for supplementary material.
\end{abstract}

\vspace{-2mm}
\keywords{Flow, Point Clouds, Tool Manipulation, Deformables} 


\section{Introduction}


\hl{Recently,} learning-based techniques \hl{have become} effective for improving the generalization capabilities of robots for manipulation tasks \hl{such} as grasping~\cite{mahler2019learning}, reorienting~\cite{openai-dactyl}, rearrangement~\cite{zeng_transporters_2020}, and tossing~\cite{zeng_tossing_2019}. 
Data observations tend to be either images~\cite{levine_finn_2016,pinto2015supersizing,rubik_cube_2019} or state information such as joint angles or end-effector poses~\cite{SAC_algos_applications_2018}, which are passed into a deep network to obtain an output vector encoding an action, typically representing a change in end-effector pose or joint angles. While these approaches have shown a wide range of successes, a fundamental limitation has to do with the nature of the observation. Using images requires projecting information into a 2D space which might lose valuable 3D information. Furthermore, learning from images in simulation often leads to a sim2real gap~\cite{reality_gap_1995} in performance. 
Although it is easy to access the internal robot states such as joint angles, the robot does not necessarily have the ground-truth state of objects in the environment, which might require complex state estimation systems~\cite{review_robot_learning_manip_2019}. Moreover, it is hard to define a state for deformable objects like liquid and cloth~\cite{manip_deformable_survey_2018,2021_survey_defs}.

In this work, we propose a framework for learning robotic manipulation from point cloud observations. Point clouds are a canonical data modality and are widely available from camera sensors, providing  valuable information about the structure of the 3D space~\cite{PointNet_2017,PointTransformer2021}. However, policy learning from point clouds has been less explored compared to learning from images or state, potentially owing to the difficulty of reasoning about raw 3D point cloud inputs. While there have been many proposed architectures which are specialized for learning from point clouds~\cite{PointNet_2017,PointNet2_2017,PointTransformer2021,VoxelNet_2017,dgcnn}, these works tend to focus on computer vision tasks such as classification and segmentation. Policy learning from point clouds, while feasible in some contexts~\cite{in_hand_manip_2021,wang2021goal}, remains challenging.

We study learning from point clouds for robotic manipulation tasks with tools. The input data is a segmented point cloud which, for each point, contains its 3D coordinates and a one-hot vector indicating the object class the point belongs to. 
Our key insight is to use \emph{dense representations} and \emph{flow} to represent the tool action. We build upon dense point-cloud processing architectures~\cite{PointNet2_2017} 
and train the model to predict per-point output values which we call \emph{tool flow}. This represents the 3D movement of each tool point in a point cloud from one time step to the next, which is an instance of scene flow~\cite{scene_flow_1999}.
Our model is trained with Behavioral Cloning on tool flow data, which provides a dense per-point supervision. Given the set of tool flow vectors, we convert flow to an SE(3) transformation, which represents the actual action a robot would execute. 
We call this model \toolflow and visualize it in Figure~\ref{fig:pull} for a pouring task in simulation.  We compare this against non-dense methods which directly regress to an action and demonstrate the benefits of tool flow as an action representation. To summarize:

\begin{figure}[t]
  \centering
  \includegraphics[width=0.98\textwidth]{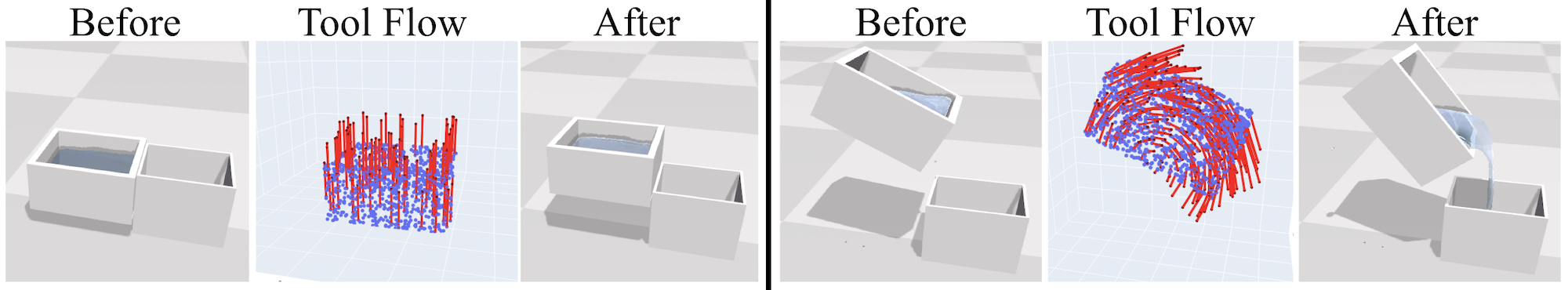}
  \caption{
  \toolflow applied on a pouring task in simulation, where the tool is the box which contains water. Given a point cloud (colored blue), \toolflow learns dense per-point flow vectors (colored red), which describe the intended 3D motion of each tool point. These are converted to translation and rotation actions. Left: the tool moves upwards. Right: the tool rotates to pour water. \hl{We subsample the flow for visual clarity}.
  }
  \vspace{-0pt}
  \label{fig:pull}
\end{figure}

\begin{itemize}[noitemsep,nolistsep]
\item We propose a general framework for learning from segmented point clouds for manipulation with tools by utilizing a novel architecture, \toolflow, which predicts per-point tool flow vectors. 
\item We show how to train \toolflow for imitation learning and explore different loss functions for training.
We perform extensive ablation studies 
to validate these choices. 
\item We perform simulated imitation learning experiments on scooping and pouring tasks and show the benefits of using \toolflow over baselines which do not use flow.
\item We demonstrate \toolflow achieves 82\% success rate on 50 physical scooping trials.
\end{itemize}


\section{Related Work}
\label{sec:rw}

\paragraph{Point Clouds and Flow}
Researchers have proposed a variety of architectures specialized for learning from point clouds~\cite{PointNet_2017,PointNet2_2017,PointTransformer2021,PointVoxelCNN,dgcnn,PointConv_2019,MeteorNet}.
We aim to explore policy learning for robotic manipulation from point clouds, and the approach we propose is compatible with any architecture producing per-point outputs from point clouds. 
Optical flow~\cite{optical_flow_1980,FlowNet2015} and its 3D counterpart, scene flow~\cite{scene_flow_1999,teed2021raft3d}, are widely used in computer vision, particularly in autonomous driving setups where the objective is to associate the movement of each pixel (or a point in 3D space) from one image (or point cloud) to the next time step. 
We use flow as an action representation for robot manipulation, and our method could integrate prior flow estimation techniques \hl{if necessary}.

\paragraph{Policy Learning from Point Clouds or Flow for Robotic Manipulation}
Learning from point clouds has been explored in grasping~\cite{wang2021goal,generalize_dexterous_manip_2022}, in-hand manipulation (by voxelizing)~\cite{in_hand_manip_2021}, visual navigation~\cite{PCL_RL_navigation_2020}, and shaping 3D deformables~\cite{partial_PC_3D_defs_2022}.
Our work differs in that we study tasks that involve manipulating a tool in 3D space from point cloud observations, and where we use tool flow as the action representation to improve learning.
While Qin~et~al.~\cite{KETO_2020} extract tool point clouds and learn keypoints for grasping and manipulating tools, we instead predict dense tool flow for manipulating the tool.
Some prior work has explored policy learning using \emph{flow} for robot manipulation, such as for fabric folding~\cite{fabricflownet}, manipulating articulated objects~\cite{articled_motions_demos_2014,FlowBot3D}, and manipulating 3D deformables~\cite{ACID2022}. This work differs in that we propose a more general framework that does not assume a specific structure of the objects being manipulated, and which predicts flow on the tool a robot controls instead of flow on a target object. Furthermore, unlike prior work~\cite{IFOR_2022} which iteratively minimizes flow with pick and place actions, or other work~\cite{dong_tactile_2021} which uses optical flow on tactile sensors,  we use flow to derive continuous tool motions in 3D space from visual input. \hl{A recent work}~\cite{FlowControl_2020} \hl{estimates optical flow using RGBD images from the current frame to the demonstration and extracts a transformation to align them. In contrast, we do not use flow for aligning frames to demonstrations but for deriving the transformations the tool should follow.}

\paragraph{Deformable Object Manipulation}
We apply our proposed tool flow framework on tasks with continuous control of a tool for deformable object manipulation. Such manipulation is challenging for robots for reasons such as the difficulty in specifying a concise state representation for deformables and their complex dynamics~\cite{manip_deformable_survey_2018,2021_survey_defs}. 
We test on scooping and pouring.
Variants of these tasks have been studied in prior work. For example,~\cite{compositional_models_2020} use scooping as an example application for task and motion planning, and~\cite{schenck_granular_2017} test scooping of granular media using a 2D image representation. Unlike these works, our approach is a general framework for robots performing continuous control of a tool to manipulate deformables in 3D space. Prior works~\cite{detecting_liquids_ISER_2016,visual_closed_loop_liquids_2017,diff_fluid_dynamics_2018,audio_flow_clarke_2018,matl_material_properties} propose methods to detect or model physics properties of granular media or liquids and test on scooping and pouring. In contrast, we propose a general method of predicting 3D tool flow which does not require modeling properties of deformables and is not specialized to scooping or pouring tasks, \hl{and which uses point clouds as inputs instead of RGB images}~\cite{DMfD_2022}.


\section{Method: \toolflow for Behavioral Cloning from Point Clouds}
\label{sec:method}

\begin{figure}[t]
  \centering
  \includegraphics[width=0.95\textwidth]{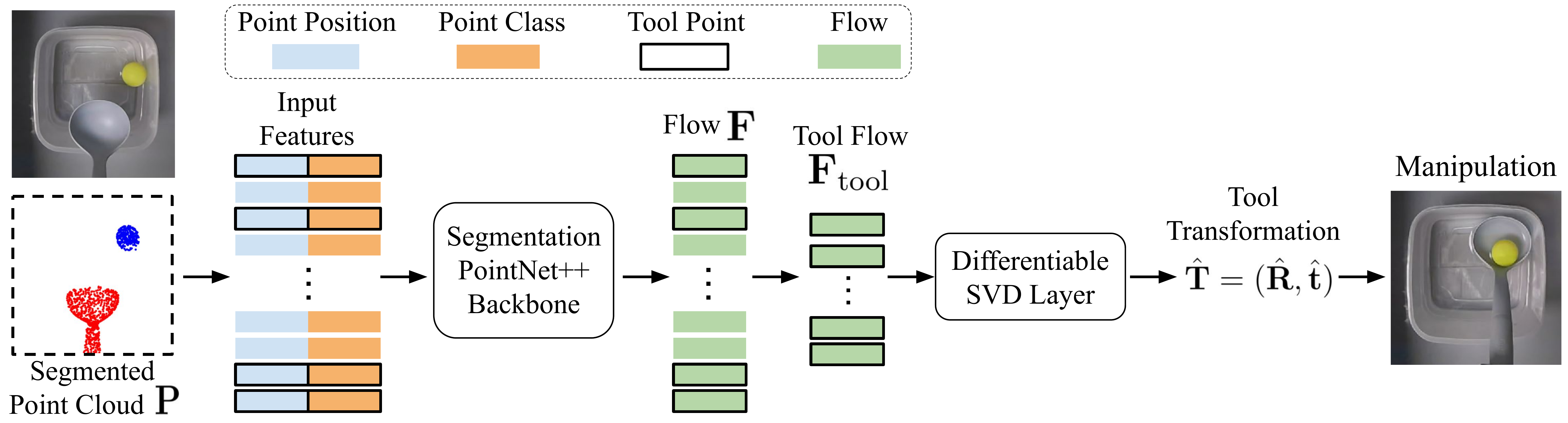}
  \caption{
  The proposed \toolflow framework learns from segmented point clouds, which form the input to a dense point cloud network to produce per-point flow vectors. We extract just the tool points (bolded above for clarity) and use those tool points to determine the transformation that the robot should apply to the tool. See Section~\ref{sec:method} for further details. Above, we show \hl{the physical scooping task}; see Section~\ref{ssec:scooping-real} for details.
  }
  \vspace{-0pt}
  \label{fig:system}
\end{figure}

We consider policy learning from segmented point cloud observations. 
A segmented point cloud $\pcl_t$ at time $t$ is an $N \times d$ array with $N$ points, each with feature dimension $d$. The feature of the $i$th point $p^{(i)} \in \pcl_t$ consists of its 3D coordinate position and a one-hot vector indicating the class of the object to which $p^{(i)}$ belongs. 
For ease of notation, we suppress the time subscript $t$ and the point index superscript $i$ when the distinction is not needed. 
We study Behavioral Cloning (BC)~\cite{Pomerleau_behavior_cloning} from segmented point clouds. BC assumes access to a dataset $\mathcal{D} = \{(\bo_t, \ba_t^*)\}_{t=1}^{M}$ of  observation-action pairs $(\bo_t,\ba_t^*)$ from a demonstrator, where $\bo_t$ indicates any type of observation  (of which segmented point clouds are one example). BC performs supervised learning to learn a policy $\pi$ with parameters $\theta$ such that the predicted action $\hat{\ba}_t = \pi_\theta(\bo_t)$ is close to the ground truth label $\ba_t^*$. While prone to compounding errors at test time~\cite{ross2011reduction}, BC has shown surprising effectiveness when compared to more complex learning-based algorithms in some robotic manipulation contexts~\cite{RB2,IBC_2021}, which motivates further study on how it can be done with point cloud observations. \hl{In future work, we will explore combining our method with other imitation learning algorithms}~\cite{Coarse_to_Fine_IL_2021,imitation_survey_2018,argall2009survey}.

\subsection{Tool Flow As Action Representation}

We propose to use \emph{tool flow} as an internal representation for the action, where the flow $f^{(i)} \in \mathbb{R}^3$ associated with point $p^{(i)}$ is a 3D vector. For a given tool point, we interpret its flow vector as representing how the point will move in 3D space as a result of ``applying'' the flow. 
To form the policy $\pi_\theta$, we use a dense point cloud network (such as a segmentation PointNet++~\cite{PointNet2_2017}), which given an input point cloud $\pcl$ produces per-point outputs.  
\hl{The point cloud input is already segmented in that it contains, for each point, the 3D world position and a one-hot encoding of its class.} 
With an $(N\times d_1)$-sized point cloud $\pcl$ as input, the output  \hl{$\flow$} has dimension $(N \times d_2)$, where $d_2$ is the output dimension (in our case, $d_2 = 3$).
We then extract \hl{from} the output \hl{$\flow$} the subset of $N' \le N$ points in $\pcl$ corresponding to all points on the \emph{tool}, while ignoring points belonging to other object classes. This results in a set of predicted 3D tool flow vectors \hl{$\flow_{\rm tool} = \{f^{(i)}\}_{i=1}^{N'}$} with $f^{(i)} \in \mathbb{R}^3$ for each tool point. 

Suppose that, at time $t$, the expert applies an action to the tool which is given by a \hl{ground-truth} transformation \hl{$\ba^* = (\bR^*, \bt^*) =$} $\bT^* \in SE(3)$.  Let $ \pcl_{\rm tool} \subseteq \pcl$ be the set of 3D points on the tool.  Then the ground-truth tool flow is given by $\flow_{\rm gt} = \bT^*  \pcl_{\rm tool} -  \pcl_{\rm tool}$ \hl{where $\bT^*  \pcl_{\rm tool}$ is the result of applying the transformation $\bT^*$ on all points in $\pcl_{\rm tool}$}. Thus, there is a one-to-one correspondence between the transformation $\bT^*$ and flow $\flow_{\rm gt}$; nonetheless, we show in Section~\ref{sec:experiments} that estimating the tool flow leads to improved performance compared to estimating the transformation $\bT^*$ directly. 

Given the set of \hl{predicted} 3D tool flow vectors \hl{$\flow_{\rm tool}$}, the next step is to extract a single overall action \hl{$\hat{\ba}$}, where \hl{$\hat{\ba}$} is a transformation that represents the change in translation and rotation of the tool's pose. 
To compute the action, we consider the tool point clouds $ \pcl_{\rm tool}$ and \hl{$ \pcl_{\rm tool}' =  \pcl_{\rm tool} + \flow_{\rm tool}$}, where in the latter, we move each point based on its estimated flow. 
Our objective is to estimate the best-fit tool transformation \hl{$\hat{\bT} = (\hat{\bR}, \hat{\bt})$} which contains rotation and translation components, respectively, to align $ \pcl_{\rm tool}$ and $ \pcl_{\rm tool}'$, \ie we want to find \hl{$\hat{\bT}$} to minimize \hl{$\| \hat{\bT} \pcl_{\rm tool} -  \pcl_{\rm tool}'\|_2$}.
To obtain the rotation \hl{$\hat{\bR}$}, we first center the two tool point clouds to obtain $\bar\pcl_{\rm tool}$ and $\bar\pcl_{\rm tool}'$.  We then input the centered point clouds to a differentiable, parameter-less Singular Value Decomposition (SVD) layer~\cite{least_squares_SVD,levinson20neurips} which computes the rotation which best aligns $\bar\pcl_{\rm tool}$ and $\bar\pcl_{\rm tool}'$ with respect to mean square error (MSE). 
The change in translation \hl{$\hat{\bt}$} can then be computed as \hl{$\hat{\bt} = C( \pcl_{\rm tool}') - \hat{\bR} C( \pcl_{\rm tool})$}, where $C(\pcl)$ denotes the centroid of the point cloud $\pcl$.
By combining the translation and rotation components, we produce the full transformation \hl{$\hat{\bT}$}, which we treat as our action representation for the policy. The outputs for the non-tool points are not supervised. We call the resulting point cloud-to-action system as \toolflow (see Figure~\ref{fig:system}), which can be used by a robot for manipulation.
\hl{Mathematically, let $F_{\theta}$ represent the segmentation PointNet++ that generates the flow vectors. ToolFlowNet computes the tool transformation as follows: }
\begin{equation}
\begin{split}
     \hat{\ba} &=  (\hat{\bR}, \hat{\bt}) = \pi_{\theta}(\bo) = \text{SVD}(\flow_{\rm tool}) \\
\end{split}
\end{equation}
\hl{where $\text{SVD}$ represents the parameter-less Singular Value Decomposition layer as described above, and $\flow_{\rm tool}$ is the flow corresponding to the tool points in the estimated flow $\flow = F_{\theta}(\pcl).$}

\begin{figure}[t]
  \centering
  \includegraphics[width=1.0\textwidth]{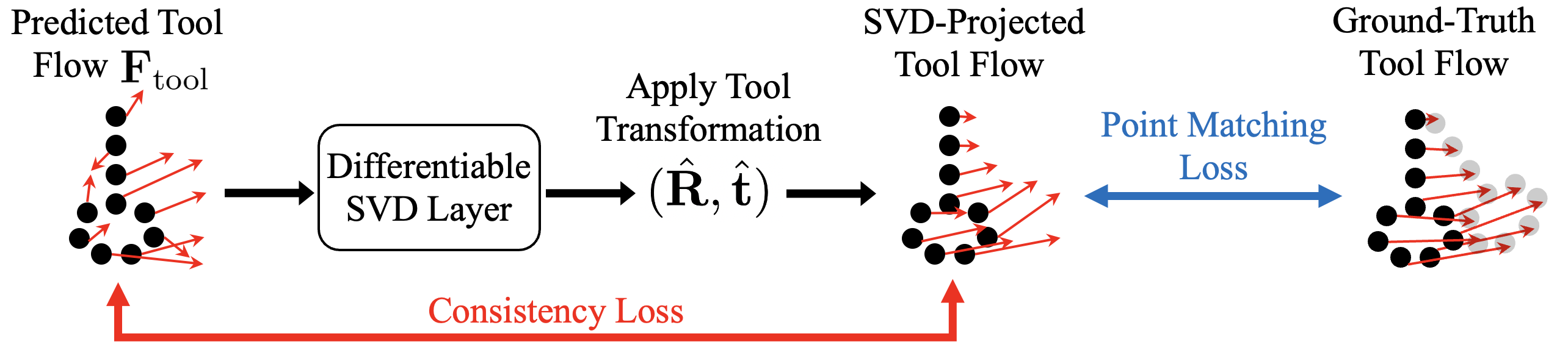}
  \caption{
  A visualization of the proposed point matching loss (Eq.~\ref{eq:point_match}) and consistency loss (Eq.~\ref{eq:consistency}). The black points visualize a simplified ladle's point cloud, and the \hl{thin} red arrows represent the flows on the tool points.}
  \vspace{-0pt}
  \label{fig:loss}
\end{figure}

\subsection{Imitation Learning via Tool Point Matching and Consistency Losses}\label{ssec:IL_losses}

\textbf{Point Matching Loss:} Given the policy's predicted action $\hat{\ba} = \pi_\theta(\bo)$, \hl{a straightforward way to imitate the ground truth action $\ba^* = (\bR^*, \bt^*)$ is to use the MSE loss: } 
\begin{equation}\label{eq:mse}
L_{\rm mse}(\hat{\ba}, \ba^*) = \beta_1 || \hat{\bR} - \bR^*||_2 + \beta_2 || \hat{\bt} - \bt^* ||_2,
\end{equation}
\hl{where $\beta_1$ and $\beta_2$ are weights for the translation and rotation components. Instead of trying to balance the weights, }
in this paper, we use a point matching loss to reduce the discrepancy between $\hat{\ba} = (\hat{\bR},\hat{\bt})$ and the ground truth action $\ba^* = (\bR^*, \bt^*)$.
Given the predicted action, the transformation $(\hat{\bR},\hat{\bt})$ 
is applied on all the original $N'$ tool points in the point cloud, and the loss function $\pointloss$ computes the Euclidean distance between the tool points transformed using the predicted action $(\hat{\bR},\hat{\bt})$ versus the tool points transformed using the ground-truth action $(\bR^*, \bt^*)$:
\begin{equation}\label{eq:point_match}
\pointloss(\pcl, \hat{\ba}, \ba^*) = \frac{1}{N'} \sum_{i=1}^{N'} \|
(\hat{\bR}p^{(i)} + \hat{\bt}) - (\bR^* p^{(i)} + \bt^*)  \|_2,
\end{equation}
where $p^{(i)}$ iterates through the $N'$ tool points in $\pcl_{\rm tool} \subseteq \pcl$, and we interpret $\hat{\bR}$ and $\bR^*$ as representing $3\times 3$ rotation matrices. Prior work on 6D pose estimation~\cite{PoseCNN,DeepIM_2018,wang2019densefusion} has used variants of this loss function to jointly optimize for translation and rotation as compared to balancing the weights on separate translation and rotation losses. 
Our usage of $\pointloss$ is similar to that in Wang~et~al.~\cite{wang2021goal} where the matching loss is on tool points directly controllable by the robot. 

\textbf{Consistency Loss:}
While $\pointloss$ should allow the policy $\pi_\theta$ to learn SE(3) pose changes (and thus, actions), its effect on optimizing the predicted flow vectors $\flow$ happens via backpropagating through a differentiable SVD layer which ``compresses'' all predicted flow vectors to produce a single transformation $(\hat{\bR},\hat{\bt})$.  This compression means that there could be significant noise in the individual flow vectors, even if they combine to form a reasonable action.
Thus, we propose a consistency loss $\consistency$ which serves as a regularizer to ensure that the \emph{predicted} flow vectors are similar to their \emph{induced, SVD-projected} flow vectors produced from the transformation encoded in $\hat{\ba}$. The loss is:
\begin{equation}\label{eq:consistency}
\consistency(\pcl, \hat{\ba}) = \frac{1}{N'} \sum_{i=1}^{N'} \|
(\hat{\bR}p^{(i)} + \hat{\bt} - p^{(i)}) - f^{(i)} \|_2,
\end{equation}
where for each of the $N'$ tool points, we compute $\hat{\bR}p^{(i)} + \hat{\bt} - p^{(i)}$ as the induced flow from the predicted transformation $(\hat{\bR},\hat{\bt})$ after applying the SVD layer, and $f^{(i)}$ is the flow predicted by the network before the SVD layer. Note that the ground truth transformation $(\bR^*, \bt^*)$ does \emph{not} appear in this consistency loss. \hl{The consistency loss is only a function of a set of points and a set of corresponding flow vectors on those points, and does not rely on any other ground truth signal.}
We combine this with the point matching loss $\pointloss$ to obtain the final loss function to optimize the policy $\pi_\theta$:
$L_{\rm combo} = \pointloss + \lambda \cdot \consistency$,
with hyperparameter $\lambda$ controlling the weight of the consistency loss, which we set to $\lambda=0.1$. See Figure~\ref{fig:loss} for visuals. \hl{To distinguish our method from traditional optical flow and scene flow methods, ToolFlowNet uses flow as a representation to compute the transformation of the tool, and is trained using the ground-truth demonstration action. It is not used to just estimate the flow.}

\textbf{Additional Implementation Details:} 
To obtain ground truth tool flow $\flow_{\rm gt}$ in simulation, we determine the 3D movement of each tool point as a result of applying the \hl{demonstrator's} action \hl{to transform those points}. 
In physical settings, we scan the tool to obtain a 3D model, from which we extract tool point clouds $\pcl_{\rm tool}$.  We perform a similar calculation where we detect the transformation executed by the robot and apply it to obtain the flow for each tool point. 
\hl{This method of extracting $\flow_{\rm gt}$ only requires access to the current observed point cloud and the corresponding action. In particular, it does \emph{not} require the perhaps more restrictive assumption of requiring one-to-one point correspondence between two consecutive point cloud observations.}
In addition, this method to detect flow means that it reflects the ``intended'' action from the robot, which may differ from the true positions of the tool points in 3D space after the robot executes the action; for example, when a collision happens with a wall, the tool points might not move, even though the robot intended for them to move. 
We leave alternative techniques to extract tool flow to future work.


\section{Experiments}
\label{sec:experiments}


\subsection{Simulation Experiments}\label{ssec:sim-tasks}

We build on top of SoftGym~\cite{corl2020softgym}, which provides a suite of deformable manipluation tasks and uses NVIDIA FleX~\cite{flex_2014} as the underlying physics engine. 
We use the simulator to obtain ground-truth segmentation labels. For the tool, we use the ``observable'' point cloud at each time step, so there may be occlusions. 
We test \hl{two} tool-based simulation tasks, \hl{PourWater and ScoopBall, and for each, test two action spaces: 3D and 6D for PourWater, and 4D and 6D for ScoopBall. In PourWater, the agent controls a box which contains water and must pour the water into a fixed target box. In ScoopBall, the agent controls a ladle and needs to scoop a ball. See Appendix}~\ref{app:sim-tasks} \hl{for more details}.

\begin{table*}[t]
  \setlength\tabcolsep{5.0pt}
  \centering
   \scriptsize
    \begin{tabular}{lcccccccc}
    \toprule
    Method & Loss & Dense & \hl{N.} Success & \hl{N.} Success & \hl{N.} Success & \hl{N.} Success & Average \\ 
    & & PN++? & \scooping \hl{4D} & \scooping \hl{6D} & \pouring \hl{3D} & \pouring \hl{6D} & \hl{N. Success} \\ 
    \midrule
    PCL Direct Vector       & MSE & \xmark & 0.544$\pm$0.03 & 0.848$\pm$0.05 & 0.530$\pm$0.08 & 0.402$\pm$0.04 & 0.581 \\  
    PCL Direct Vector          & PM  & \xmark & 0.228$\pm$0.12 & \hl{0.048$\pm$0.04}  & 0.132$\pm$0.07 & \hl{0.088$\pm$0.04} & \hl{0.124} \\
    PCL Dense Transformation   & MSE & \cmark & 0.519$\pm$0.07 & 0.824$\pm$0.06 & 0.539$\pm$0.05 & 0.344$\pm$0.03 & 0.556 \\  
    PCL Dense Transformation   & PM  & \cmark & 0.367$\pm$0.07 & \hl{0.360$\pm$0.10}  & 0.583$\pm$0.03 & \hl{0.049$\pm$0.02}  & \hl{0.340} \\
    \hl{D Direct Vector}            & MSE & \xmark & \hl{0.190$\pm$0.07} & \hl{\textbf{0.952$\pm$0.02}} &  \hl{0.035$\pm$0.01} & \hl{0.069$\pm$0.02} & \hl{0.311} \\
    \hl{D+S Direct Vector}          & MSE & \xmark & \hl{0.734$\pm$0.11} & \hl{\textbf{0.928$\pm$0.03}}  & \hl{\textbf{0.777$\pm$0.03}} & \hl{0.304$\pm$0.03}  & \hl{0.686} \\
    \hl{RGB Direct Vector}          & MSE & \xmark & 0.354$\pm$0.05 & \hl{0.776$\pm$0.05}  & 0.698$\pm$0.02 & \hl{0.324$\pm$0.05}  & \hl{0.538} \\
    \hl{RGB+S Direct Vector}        & MSE & \xmark & \hl{0.671$\pm$0.07} & \hl{\textbf{0.944$\pm$0.02}} & \hl{\textbf{0.804$\pm$0.04}} & \hl{0.353$\pm$0.03} & \hl{0.693} \\
    \hl{RGBD Direct Vector}         & MSE & \xmark & \hl{0.418$\pm$0.10} & \hl{0.920$\pm$0.02} & \hl{0.733$\pm$0.07} & \hl{0.353$\pm$0.02} & \hl{0.606} \\
    \hl{RGBD+S Direct Vector}       & MSE & \xmark & \hl{0.734$\pm$0.10} & \hl{\textbf{0.968$\pm$0.02}} & \hl{\textbf{0.830$\pm$0.03}} & \hl{0.481$\pm$0.03} & \hl{0.753} \\
    \midrule
    \toolflow, No Skip Conn.  & PM+C  & \cmark & 0.987$\pm$0.08 & \hl{0.304$\pm$0.06}  & 0.000$\pm$0.00 & \hl{0.000$\pm$0.00} & \hl{0.323} \\
    \toolflow, MSE after SVD  & MSE+C & \cmark & 0.089$\pm$0.04 & \hl{0.792$\pm$0.09}  & 0.494$\pm$0.02 & \hl{\textbf{0.913$\pm$0.05}} & \hl{0.572} \\
    \toolflow, PM before SVD  & PM  & \cmark & 0.785$\pm$0.08 & \hl{0.880$\pm$0.05}  & 0.618$\pm$0.04 & \hl{0.677$\pm$0.05} & \hl{0.740} \\
    \toolflow, No Consistency & PM  & \cmark & 0.861$\pm$0.06 & \hl{0.744$\pm$0.12}  & 0.468$\pm$0.10 & \hl{0.609$\pm$0.06} & \hl{0.670} \\
    \midrule
    \textbf{\toolflow (Ours)} & PM+C  & \cmark & \textbf{1.152$\pm$0.07} & \hl{\textbf{0.952$\pm$0.02}} & \textbf{0.795$\pm$0.05} & \hl{0.667$\pm$0.03} & \hl{\textbf{0.892}} \\
    \bottomrule
    \end{tabular}
  \caption{
  Behavioral Cloning (BC) results in simulation. The first \hl{10} rows are baselines, the next 4 are ablations of our method, and the last row is our method. We report the loss functions used \hl{as MSE only, PM only (the loss in Eq.}~\ref{eq:point_match}\hl{), or if it also uses a consistency loss (+C, from Eq.}~\ref{eq:consistency}\hl{).} \hl{We also show} whether the method uses a segmentation PointNet++ (i.e., a dense architecture), and the \emph{normalized} success rates \hl{(N. Success)} across all tasks over 5 independent BC runs. The last column averages the success across the four columns.
  \hl{We bold the best numbers in the columns, plus any with overlapping standard errors.} 
  }
  \vspace*{-0pt}
  \label{tab:bc-results-stderr-v2}
\end{table*}

\subsubsection{Baseline Methods}
\label{ssec:baseline_methods}

We compare the proposed method with the following baselines (see Section~\ref{ssec:ablations} for ablations):

\begin{itemize}[noitemsep,nolistsep]
\item \textbf{\hl{PCL} Direct Vector}. Uses a \emph{classification} PointNet++ network to directly estimate a vector action (with a translation and an axis-angle rotation). \hl{We test two variants, one which supervises with the MSE loss and another which uses the Point Matching (PM) loss from Eq.}~\ref{eq:point_match} \hl{on tool points.}
\item \textbf{\hl{PCL} Dense Transformation}. Uses a \emph{segmentation} PointNet++, and directly predicts per-point 6D vectors (translation and axis-angle). Each point cloud has a designated point as the center of rotation for the tool, and we use the output corresponding to that point as the vector action. The outputs for the other points are not supervised. This baseline is designed to isolate any benefits from using the segmentation version of PointNet++ instead of classification. \hl{As with Direct Vector, we test two variants, with supervising using the MSE or PM losses}. 
\item \hl{\textbf{\{D, RGB, RGBD\} Direct Vector}. Processes images and uses a Convolutional Neural Network to directly predict an action vector (translation and axis-angle) and supervises with MSE. The inputs are either a depth image (D), the RGB image, or an RGBD image.} 
\item \hl{\textbf{\{D+S, RGB+S, RGBD+S\} Direct Vector}. These are the same as the prior set of methods, except that the input images have additional channels corresponding to binary segmentation masks. We denote these new input images as: D+S, RGB+S, and RGBD+S. We include these baselines for a fairer comparison due to assuming segmentation information in the point cloud observations.}
\end{itemize}

\subsubsection{Experiment Protocol and Evaluation}\label{ssec:exp-protocol}

For each task, we use \hl{a scripted} demonstrator to generate a fixed set of training demos and keep the successful ones for Behavioral Cloning. 
We standardize on \hl{500 training} epochs for all methods \hl{and average across 5 seeds}. We evaluate every \hl{25} training epochs on \hl{25} testing configurations and use the maximum success (averaged over 5 seeds) across the full training history, then divide this by the demonstrator success rate to get the normalized performance.
\hl{See Appendix}~\ref{app:evaluation} \hl{for more details.}

\subsection{Simulation Results and Analysis}
\label{sec:results}

\begin{figure}[t]
  \centering
  \includegraphics[width=0.99\textwidth]{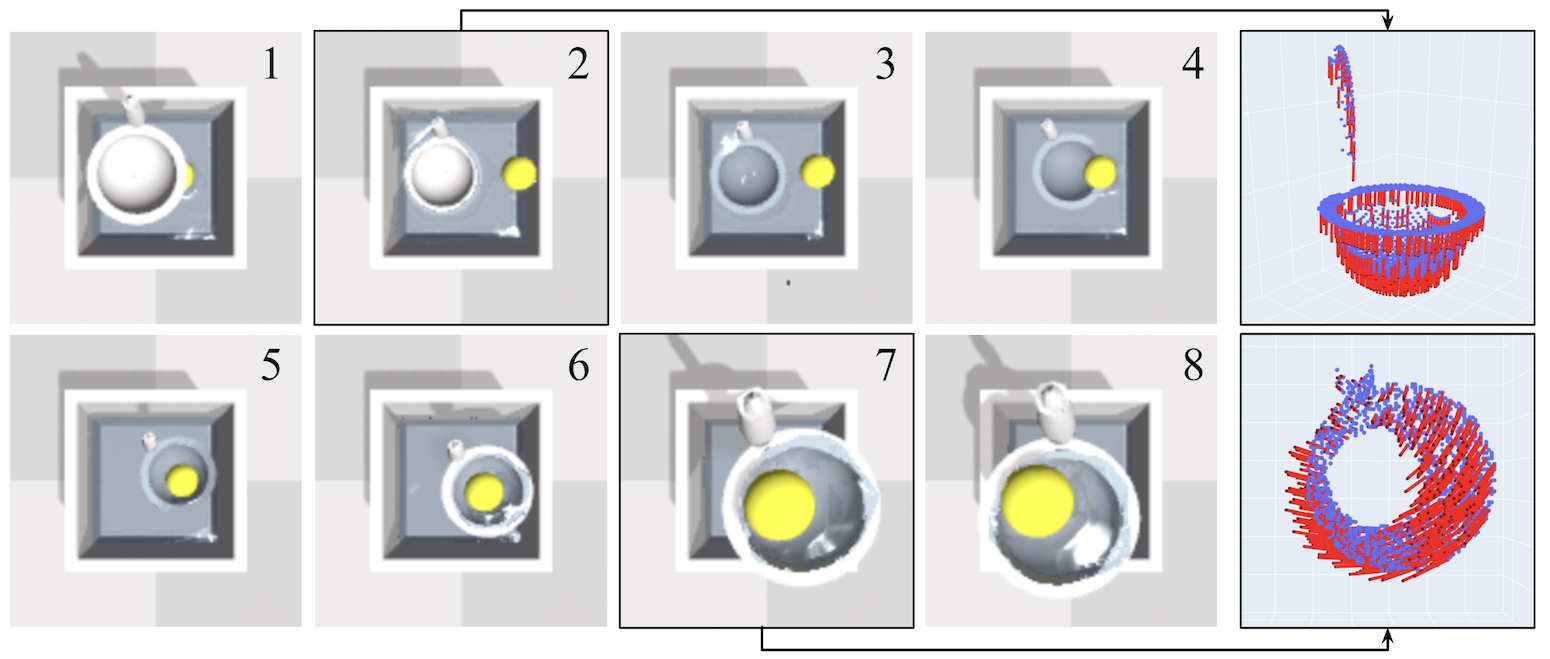}
  \caption{
  An example successful \scooping \hl{4D} rollout by a learned \toolflow \hl{Behavioral Cloning policy}. We show subsampled RGB frames for visual clarity, though the policy only uses point clouds as input. For two of the frames, we show the policy's tool flow visualizations. The policy lowers the ladle (frames 1-3), rotates and moves it in the direction of the ball (frames 4-5), lifts the ball (frame 6) and then rotates back to the starting pose (frames 7-8). The policy's flow visualizations for frames 2 and 7 suggest the ability to learn downward and rotation movement, respectively. The predicted flow vectors, colored red, are slightly enlarged for clarity.
  }
  \vspace{-0pt}
  \label{fig:results-mm-1}
\end{figure}

The results in Table~\ref{tab:bc-results-stderr-v2} suggest that using \toolflow outperforms the baselines on average across the tasks. In particular, for ScoopBall 4D and PourWater 6D, it outperforms all other baselines, and for ScoopBall 6D and PourWater 3D, it is on par with the best image-based baselines. This may indicate that some tasks have a 3D nature which makes it more natural to learn policies from point clouds. 
Figure~\ref{fig:results-mm-1} shows a qualitative example test-time rollout of \scooping 4D from the learned \toolflow policy. 
Figure~\ref{fig:results-mm-1} also visualizes the policy's internal flow predictions (\ie the per-point flow vectors $f^{(i)}$ before the SVD layer), showing that the network has learned surprisingly clean per-point tool flow vectors. Furthermore, as the agent controls the ladle at its upper tip, when rotating, the flow vectors also correctly predict longer flow vectors for the points located further away from the origin of the tool pose.

\subsection{Why Does \toolflow Help Robot Learning?}\label{ssec:ablations}

We perform further experiments to determine why \toolflow outperforms the baselines that directly regress to a transformation.  Specifically, we create a variant of \scooping in which the action space consists of translations only (no rotations); see Appendix~\ref{app:why-tool-flow-helps} for details. These experiments reveal that \toolflow does not outperform the baselines in translation-only settings, indicating that the benefits of \toolflow come from predicting rotations. We also test Direct Vector methods with 4D (quaternions), 6D~\cite{Continuity_Rotations_2019}, 9D (rotation matrices)~\cite{levinson20neurips}, and 10D~\cite{rotations_10D_rss_2020} rotation representations in Appendix~\ref{app:rpmg-rotations}, and find that \toolflow continues to obtain higher success rates.


We next study ablations of \toolflow to understand which components are most critical:

\begin{itemize}[noitemsep,nolistsep]
\item \textbf{\toolflow, No Skip Connections}: removes skip connections in the segmentation PointNet++.
\item \textbf{\toolflow, MSE after SVD}: tests applying an MSE loss on the induced transformation from SVD instead of point matching. We still use the consistency loss (Eq.~\ref{eq:consistency}).
\item \textbf{\toolflow, Point Matching (PM) Before SVD}: tests using the PM loss (Eq.~\ref{eq:point_match}) before the SVD layer, so the loss does not back-propagate through the SVD layer.
\item \textbf{\toolflow, No Consistency}: tests removing the consistency loss (and just using Eq.~\ref{eq:point_match}).
\end{itemize}

We use the same experiment protocol as in Section~\ref{ssec:exp-protocol} on all tasks. The results, also in Table~\ref{tab:bc-results-stderr-v2}, suggest strong benefits to using the point matching loss, the consistency loss, and the standard segmentation PointNet++ with skip connections. For example, \hl{across all tasks, ToolFlowNet} performance is worse without using a consistency loss. The utility of some design choices may be more task-specific; removing skip connections leads to no successes on \pouring because removing it made the model unable to predict any rotations \hl{(see Appendix}~\ref{app:main-experimental-results} \hl{for additional analysis)}, while it is possible to succeed in \scooping without using rotations.

\subsection{Physical Scooping Experiments}
\label{ssec:scooping-real}

\begin{figure}[t]
  \centering
  \includegraphics[width=0.99\textwidth]{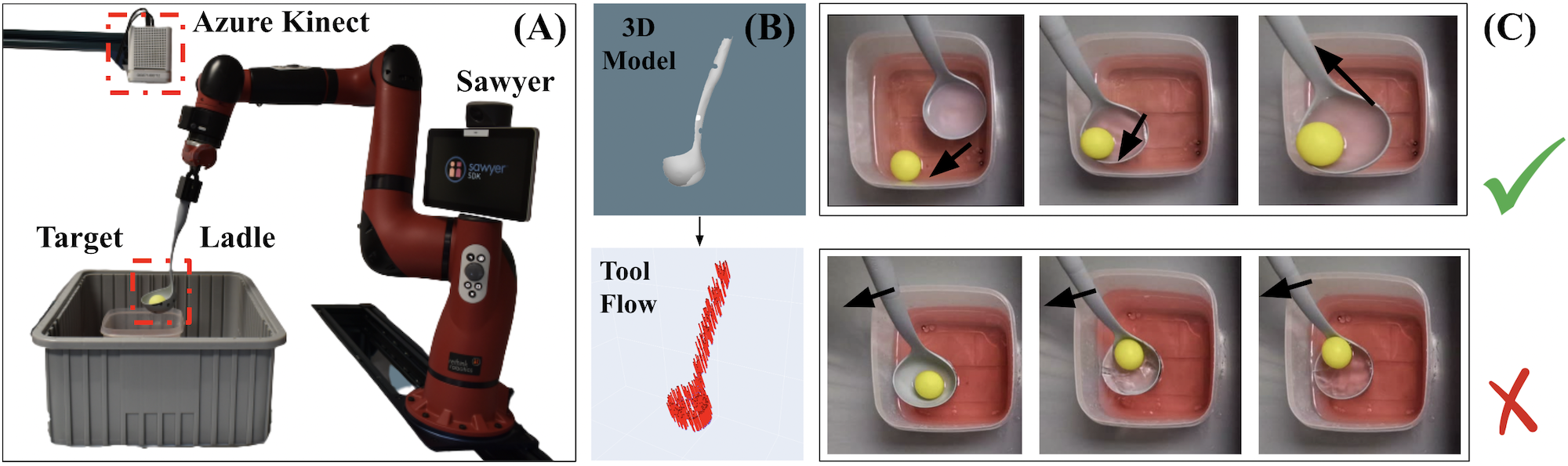}
  \caption{
  Physical experiments. (A) The Sawyer holds a ladle above a small box with water, which is enclosed in a larger gray box to contain spills. (B) The scanned 3D model of the ladle with a representative tool flow visualization. (C) Example test-time trials with subsampled frames. Top row: successful tool movement towards and lifting the target. Bottom row: collision failure due to repeatedly pushing against the container.
  }
  \vspace{-0pt}
  \label{fig:physical}
\end{figure}

We test \toolflow in the real world using a Sawyer robot with a standard consumer ladle which we scan to obtain a 3D model, and a yellow ping-pong ball acting as the target item (see Figure~\ref{fig:physical}). The ladle is attached to the Sawyer's end-effector. As in simulation, we obtain tool flow by recording the change in end-effector pose and applying the transformations on the tool point cloud. A Microsoft Azure Kinect camera captures top-down depth images to compute the ball's point cloud.

\begin{wraptable}{r}{4.0cm}  
  \vspace{-5pt} 
  \scriptsize  
  \begin{tabular}{lr}
  \toprule
  \toolflow in Real & \#Trials \\ 
  \midrule
  Successes & \hl{41}/50   \\
  Failures  & \hl{9}/50   \\
  \bottomrule
  \end{tabular}
  \caption{
  \scriptsize
  Physical scooping results.
  }
  \vspace{-10pt}
  \label{tab:physical}
\end{wraptable}
A human operator manually moves the Sawyer's arm via direct touch to collect \hl{125} demonstrations, with each comprising about \hl{20} observation-action pairs. We use \hl{100} demonstrations for training \toolflow, and use the remaining \hl{25} for monitoring evaluation MSE. We perform 50 physical scooping trials, where each trial begins with the human dropping the ping-pong ball \hl{over the} water \hl{at an arbitrary location within the inner box shown in} Figure~\ref{fig:physical}. The trial is \hl{classified as} successful if the Sawyer raises the ball from the water surface to \hl{above the} top of the smaller box. Results in Table~\ref{tab:physical} suggest that the Sawyer achieves \hl{41}/50 successes (82\%), with \hl{9} failures. All failures were due to the ladle colliding with the small box. See Appendix~\ref{app:physical} for more details. In future work we will do physical experiments with more complex demonstrations.

\subsection{Limitations and Failure Cases}

In our experiments, we obtain the ground-truth tool flow data by applying the demonstrator's actions on a set of tool points  and computing the change in the resulting tool point positions. \hl{The tool points can be observed or derived via a tool model. In either case, we} require access to the demonstrator's action, and future work could relax this assumption by extracting tool flow without explicit actions, such as when a human provides a video. Possible approaches include using scene flow techniques.

A limitation of \toolflow \hl{is that it may be susceptible to occlusions of the tool when a model of the tool is not available}. For physical scooping, we \hl{rely on a scanned model of the tool because the Sawyer's arm would occlude parts of the tool, but tool models might not always be available}. In future work, we will explore ways to address occlusions such as point cloud inpainting and tracking. 
Finally, we test \toolflow on two simulation tasks \hl{with two action spaces for each,} and scooping in real. We hope to test on more diverse tasks such as cloth or rope manipulation, and to address failures from the physical experiments.


\section{Conclusion}
\label{sec:conclusion}

In this work, we propose a general technique for policy learning from point clouds for tool-based manipulation tasks, which we demonstrate on scooping and pouring tasks. Our method, called \toolflow, predicts per-point tool flow vectors which are converted into actions. We hope this research inspires future work on learning from point clouds.



\acknowledgments{
This work was supported by LG Electronics and by NSF CAREER grant IIS-2046491. We thank Brian Okorn and Chuer Pan for assistance with the differentiable SVD layer, and Mansi Agrawal, Sashank Tirumala, and Thomas Weng for paper writing feedback.
}


{\footnotesize
\bibliography{example}  
}

\clearpage

\appendix
\beginsupplement 
\etocsetlevel{appendixplaceholder}{-1}
\etoctoccontentsline*{appendixplaceholder}{APPENDIX}{-1}
\localtableofcontents

\clearpage

\section{Implementation Details}

\subsection{The Simulation Tasks}\label{app:sim-tasks}

In this section, we discuss in further detail the \scooping and \pouring simulation tasks we introduced in Section~\ref{ssec:sim-tasks}.

\subsubsection{Shared Properties of Simulation Tasks} 

For \hl{both} tasks, point cloud observations have a maximum of $N=2000$ points, and there may be fewer points at certain time steps. We do not zero-pad the point clouds to keep them a fixed data dimension. 
To obtain segmented point clouds, we use the depth images from SoftGym and project each pixel into world coordinates to create a point cloud.  Since we use the observable point cloud at each time step for the tool, there may be occlusions, \ie if certain parts of the tool are not visible to the depth camera, they will not be included in the point cloud. For example, the second flow visualization in Figure~\ref{fig:results-mm-1} shows that the ball occludes part of the point cloud.

\hl{Both} tasks use a fixed episode length of 100 time steps (there is no early termination) along with an action repeat of 8, meaning that each action $\ba$ from the policy or demonstrator is executed 8 times ``internally'' which corresponds to 1 of the 100 time steps.
\hl{For further investigation, we test each task using two different action spaces.}
Each task and action space comes with a scripted demonstrator. Videos of the demonstrator for all tasks are available on the project website: \url{https://tinyurl.com/toolflownet}.

The simulation is built upon FleX~\cite{flex_2014}, which is a particle-based simulator. Hence, we sometimes may use ``water particles'' or ``ball particles'' to describe the physics of those items. The tools for both tasks are not particle-based. In FleX simulation, when the tools move, they affect the position of the particles (but not vice versa).

\subsubsection{\scooping, Details of the Task and Demonstrator}\label{app:scooping-more-details}

\noindent \textbf{Overview}.
The agent controls a ladle, and each episode begins with the ladle above a box that contains water and a single ball floating on it. The objective is to scoop the ball above a certain height. 
Each episode has a different initial position of the ball.
\hl{There are 2 versions of the task, with 4 DoF and 6 DoF actions. For the former,} the action space $\ba = (\Delta x, \Delta y, \Delta z, \Delta \theta)$ is 4D, which consists of the change in the coordinates of the ladle tip, and the change in rotation about the vertical axis $\Delta \theta$ coinciding with the ladle's tip. See Figure~\ref{fig:results-mm-1} for visuals of the translation and rotation actions. We ignore the unused 2 dimensions from the 6D output transformation of \toolflow to get a 4D action. \hl{For the 6 DoF action version, the action is 6D and consists of 3D delta translation and 3D delta rotation (which is the 3D delta axis-angle in the local tool frame). We also add a hole in the ladle for the 6 DoF action version to let water leak from it, which significantly stabilizes the water-ball simulator dynamics. }


\noindent \textbf{Point Cloud}.  
The point cloud uses two classes, corresponding to the tool (\ie ladle) and the target ball. The water particles are not part of the point cloud, as knowledge of the water particles is not critical to succeed at the task. The task uses the observable point cloud for both the tool and the ball, so the tool can occlude the ball (and vice versa). The water particles do not occlude the tool. 

\noindent \textbf{Success Criteria}.  
A \hl{binary} success is triggered when the agent keeps the ball above a height threshold (above the water height) for at least 10 time steps, which ensures that the ball must be at a reasonably stable state for a success. The agent can only do this by using its ladle to scoop the ball. 

\noindent \textbf{Physics Considerations}.  
We create a ladle model in Blender\footnote{\url{https://www.blender.org/}} and import the resulting model into SoftGym as a signed-distance function. Since the FleX physics backend does not provide collision checking for arbitrary meshes, we implement our own approximate version. Whenever the agent applies an action, we compute the sphere formed from ``completing'' the ladle's bowl, and then compute whether it has intersected with the walls of the box. If an intersection exists, then we clip each action dimension such that the resulting state is still legal (\ie respects collision boundaries).

The physics of water-ball interaction in FleX have a great impact on this task. We set the ball so that it has a density that should make it always float on water. However, when the agent scoops the ball, the water particles may easily ``push'' the ball away, causing the ball to fall back into the water. (This behavior does not happen in our physical experiments.) Furthermore, when the ball drops from midair into the water, sometimes the ball remains sunk afterwards, making it impossible for the agent to succeed with the given action space and ladle physics. In future work, we will investigate simulators that have improved liquid-solid physics interactions. \hl{For this task, we originally began tests with a 4 DoF action space which used a ladle with a solid bowl, which meant during scooping that its water paticles could push away the ball. The 6 DoF action version, however, uses a ladle with a hole in it to let water drain, which significantly improved success rates.}

\noindent \textbf{Demonstrator}.  
\hl{With 4 DoF actions,} we implement an algorithmic demonstrator which first lowers the ladle into the water, attempts to move the ladle so that its bowl is underneath the ball, then lifts the ball. The demonstrator continually rotates the ladle so that the ladle's handle is facing the direction of the ball. When the demonstrator lifts, it rotates again so that the ladle is back at the starting rotation. The process of lowering and lifting the ladle results in y-coordinate\footnote{In SoftGym, the positive y-axis points upwards, while the x- and z-axes form a flat horizontal plane.} changes of 0.004 in SoftGym simulation units, while translations within the water results in actions similarly bounded by 0.004 units in both coordinate directions. When translations get scaled by 250X (see Appendix~\ref{app:scaling-targets}), the targets have per-component magnitude upper bounded by $0.004 \times 250 = 1.0$. Each rotation consists of a change in 0.5 degrees about the axis coinciding to the ladle's ``stick.'' Largely owing to the aforementioned water-ball simulation artifacts, the demonstrator success rate is 63.2\%. \hl{We filter the resulting data to only imitate successful demonstrations}.

\hl{With the 6 DoF version with the different ladle, we re-script the demonstrator to execute a more visually natural scoop which uses all its rotations, and then moves towards the ball, then rotates back to a neutral position and lifts upwards. The demonstrator success rate here is 100.0\%.}

\subsubsection{\pouring, Details of the Task and Demonstrator}

\noindent \textbf{Overview}. 
We use the \pouring task from SoftGym~\cite{corl2020softgym}. \hl{The task comes with a 3 DoF action space, so to explore more complex action spaces, we modify the code to support a 6 DoF action space.} 
The agent controls a box which contains water and must pour the water into a fixed target box.
\hl{There are two versions of the task, with 3 DoF actions and 6 DoF actions. The 3 DoF action space $\ba = (\Delta x, \Delta y, \Delta \theta)$ consists of the change in the $x$ and $y$ coordinates of the controlled box's center and the change in rotation $\Delta \theta$ around the bottom center of the box along a single coordinate axis.  See Figure}~\ref{fig:pull} \hl{(left) for the translation and Figure}~\ref{fig:pull} \hl{(right) for the rotation. For the 6 DoF version, the action is 6D and consists of the translation change in all $x, y, z$ coordinates, and rotational change around the bottom center of the box along all 3 coordinate axes. The rotation is expressed as Euler angles represented as an extrinsic rotation about the Z-Y-X axes in that order. The rotational change is a delta in the Euler angles for all axes.} 
In each episode, we vary the sizes of both boxes, the amount of water in the controlled box, and the starting distance between the boxes. \hl{For the 6 DoF action version, we also vary the initial orientation of the controlled box. }
The proposed \toolflow generates full 6D transformations, and we ignore the unused 3 action dimensions at test time \hl{for the 3 DoF action version}.

To make the task more tractable, we slightly reduce the maximum possible box to be about 75\% of the maximum size compared to the public version. \hl{For the 6 DoF action space, we add more randomness to the initial starting pose of the controlled box}.  Other than that, \hl{for the 3 DoF action space,} we keep the \pouring settings as consistent with the open-source code as possible to potentially facilitate comparisons with other work using this task.

\noindent \textbf{Point Cloud}.  
The point cloud uses three classes, corresponding to the tool (\ie the box that starts with water), the target box for the water, and the water itself. Here, we use the observable point cloud for the two boxes, but for the water, we follow the existing SoftGym implementation and use the ground truth water particle positions which we can query at each time step. We include the water in the point cloud because knowledge of the water is essential for pouring.

\noindent \textbf{Success Criteria}.  
We set a binary success threshold based on if at least 75\% of the water particles end in the target box.

\noindent \textbf{Physics Considerations}.  
One of the FleX simulation artifacts is that water particles can ``seep through'' the corners and edges of both boxes. The 75\% particle threshold we use is high enough to convey reasonable task success, but not so high that it cannot tolerate some water particles escaping from the boxes.
To handle collisions, for the 3 DoF action space, we use the existing collision checking code from SoftGym without further modification. If the tool intersects with the bottom floor, or intersects with the target box, the action is not applied, which can cause the tool to ``freeze'' if repeatedly applying collision-violating actions. We implement a similar collision checking code for the new 6 DoF action space.

\noindent \textbf{Demonstrator}.
\hl{For both action versions, we script a demonstrator which moves the box towards the target and rotates to pour the water.}
\hl{With 3 DoF actions,} we implement an algorithmic demonstrator which moves the box towards the target, lifts the box, then rotates to pour the water in the target. The act of moving towards the box consists of translations in the positive x direction of 0.003 units, lifting consists of translations in the positive y direction of 0.003 units, and then rotating consists of rotating 0.5 degrees in the positive direction to pour, and then rotating negative 0.5 degrees to reset back to the original orientation. When translations get scaled by 250X, this creates targets with per-component magnitudes of $0.003 \times 250 = 0.75$. The demonstrator success rate is 90.6\%.

\hl{We script a similar demonstrator for the 6 DoF action version. The controlled box starts from a more complex configuration, so the demonstrator rotates and translates it to align it with the target. Then, it does a similar maneuver as the 3 DoF demonstrator to pour the box with water into the target box. Its success rate is 81.5\%.}

\subsection{More Details on Simulation Experiments}

We present more details of the simulation experiments reported in Section~\ref{sec:experiments} (and in Appendix~\ref{app:experiments}).

\subsubsection{Training Hyperparameters}

\begin{table*}[h]
  \setlength\tabcolsep{6.0pt}
  \vspace*{-0pt}
  \centering
  \footnotesize
    \begin{tabular}{lccrr}
    \toprule
    Network    & Epochs & LR & Batch & Params \\ 
    \midrule
    PointNet++ & 500 & 1e-4 &  24 & 1.4M \\
    CNN        & 500 & 1e-4 & 128 & 3.0M \\
    \bottomrule
    \end{tabular}
  \caption{
  Some hyperparameters used in experiments. We report the number of training epochs, the Adam learning rate, the batch size, and the number of network parameters.
  }
  \vspace*{-0pt}
  \label{tab:hyperparams}
\end{table*}  

For a representative set of hyperparameters, see Table~\ref{tab:hyperparams}. We train models for the same number of epochs with a common Adam~\cite{adam2015} learning rate of 1e-4. However, the CNN uses a larger batch size and has more than 2X as many parameters as compared to the PointNet++. Here, we use PointNet++ to refer to both the segmentation version (as used in \toolflow) and the classification version (used in some baselines), which have almost the same number of parameters (1.4M).

\subsubsection{Experiment Protocol and Evaluation Metrics}\label{app:evaluation}

\hl{For each task, we generate a set of starting configurations and divide them into training and testing configurations. Each configuration has a slightly different arrangement of particles, so that the policies cannot succeed by memorizing the training data. We use an algorithmic demonstrator (described in Appendix}~\ref{app:sim-tasks} \hl{) to generate a fixed set of training demonstrations from the starting training configurations. Following prior work in imitation learning}~\cite{seita_bags_2021}\hl{, we filter the demonstrations to keep only the successful ones for Behavioral Cloning. For a fair comparison, all comparisons among methods on a single task and action space train on the same set of demonstrations.}


\hl{For simulation experiments, we standardize on 100 training demonstrations for all tasks except for ScoopBall 6D, for which we use 25 training demonstrations. See Appendix}~\ref{app:num-demos} \hl{for experiments where we adjust the number of training demonstrations.}

We evaluate Behavioral Cloning performance by training for 500 epochs on task-specific demonstration data. For each method, we perform 5 independent runs (each with a different random seed), and evaluate every 25 epochs on 25 fixed held-out starting configurations. In other words, we test ``snapshots'' of each training run every 25 epochs. As all episodes have a binary success outcome, averaging the 25 test episodes gives us one quantitative number for each epoch in a training run. We then average over the 5 Behavioral Cloning runs for each epoch, and treat that number as the method's performance at each epoch. Thus, each epoch's resulting metric reflects $25\times 5 = 125$ total evaluation rollouts, and we then consider the maximum over all the epochs, since in Behavioral Cloning we often just care about the best snapshot at any time (due to no environment interaction). That provides a number corresponding to raw success rate. We finally normalize by dividing this value by the demonstrator performance, which gives us the final normalized success rate.

Due to noise and variance in the learning process~\cite{DeepRLMatters}, for a given method, we report not just the average normalized performance but also the corresponding standard error of the mean, which here is the sample standard deviation divided by $\sqrt{5}$. In addition, \textbf{when bolding numbers in tables} to indicate the ``best'' method, we bold both the best number \emph{and} those that have overlapping standard errors. For example, when comparing just $x_1 \pm y_1$ versus $x_2 \pm y_2$, if $x_1 > x_2$, then we would bold the $x_1$ value in a table, \emph{along with} $x_2$ if the condition $x_2 + y_2 \ge x_1 - y_1$ holds.

In Appendix~\ref{app:main-experimental-results}, we report an alternative evaluation metric where we instead take an average over all epochs instead of picking the best one.

\subsubsection{Implementation of Baseline Methods}

To keep experimental settings fair, we strive to apply as consistent settings as possible among the methods. For example, \hl{if we evaluate using an action space with fewer than 6 DoFs,} we perform the same procedure to convert a 6D action prediction into a 3D action (for \pouring) or a 4D action (for \scooping) for all methods by zeroing out unused action dimensions.
In addition, the baselines that use the classification PointNet++ architecture, \textbf{Direct Vector} with either the MSE (Equation~\ref{eq:mse}) or Point Matching (Equation~\ref{eq:point_match}) losses, use an architecture with roughly the same amount of parameters (approximately 1.4M) as the segmentation PointNet++ architecture.

We test with two baselines we call \textbf{Dense Transformation} with (again) variants based on using the MSE or Point Matching losses. For these methods, we pick one fixed point on the tool and use that to represent the point of interest. The segmentation PointNet++ produces per-point outputs, so this fixed point tells us which one, out of all the output points, provides the transformation (\ie action). For \scooping we use the tip of the ladle, and for \pouring we use the center of the bottom of the box. These both coincide with the center of the tool rotation, and which we treat as synthetic points in the tool point cloud. We extract them using ground truth simulator knowledge (instead of the observable point cloud, since they may be occluded or out of view) and insert them into the segmented point cloud $\pcl$. These form one point on the tool; the point cloud still contains the usual amount of tool points in the observable point cloud.

\hl{For methods that process images, we use a Convolutional Neural Network (CNN) to process the images.} We use an architecture similar to the design of the CNN encoder in the SAC/CURL code repository~\cite{CURL}, but where we slightly reduce the parameter count to be about 3M, which is still more than the 1.4M parameters for the PointNet++ models. In PyTorch print string format, assuming a three-channel image input, the network is expressed as:

{\footnotesize
\begin{verbatim}
Actor(
  (encoder): PixelEncoder(
    (convs): ModuleList(
      (0): Conv2d(3, 16, kernel_size=(3, 3), stride=(2, 2))
      (1): Conv2d(16, 16, kernel_size=(3, 3), stride=(1, 1))
      (2): Conv2d(16, 16, kernel_size=(3, 3), stride=(1, 1))
      (3): Conv2d(16, 16, kernel_size=(3, 3), stride=(1, 1))
    )
    (fc): Linear(in_features=29584, out_features=100, bias=True)
    (ln): LayerNorm((100,), eps=1e-05, elementwise_affine=True)
  )
  (trunk): Sequential(
    (0): Linear(in_features=100, out_features=256, bias=True)
    (1): ReLU()
    (2): Linear(in_features=256, out_features=256, bias=True)
    (3): ReLU()
    (4): Linear(in_features=256, out_features=6, bias=True)
  )
) 
\end{verbatim}
}

\hl{The RGB and depth input images have resolution 100x100. When testing with RGB and depth (i.e., RGBD) images, we stack the images channel-wise. We also augment these baseline methods to include binary segmentation image masks as extra image channels. This is designed to reproduce the same segmentation information that is present in a segmented point cloud. With ScoopBall, there are two binary segmentation mask images, one for the tool and one for the ball. If testing using RGB image inputs with segmentation masks (denoted as RGB+S in tables), for example, then this results in 5-channel input images to the CNN, with the first three for RGB and the last two for the segmentation masks. PourWater, however, has three binary segmentation masks, corresponding to the controlled cup, the target cup, and the water. Thus, for PourWater, RGB+S image inputs are 6-channel images.}

For consistency, the CNN architecture we use is the same across all methods that process image inputs.

\subsubsection{Implementation of \toolflow}

For the policy architecture $\pi_\theta$, we build on the PointNet++ implementation from PyTorch Geometric~\cite{pytorch_geometric}. We keep the architecture and hyperparameters similar to those in PyTorch Geometric. Both the segmentation and classification PointNet++ use two Set Abstraction levels~\cite{PointNet2_2017} with ratio parameters 0.5 and 0.25 and ball radius parameters 0.2 and 0.4 for the two layers, respectively. These two Set Abstraction levels are then followed by a third ``Global'' Set Abstraction layer which performs a global max-pooling operation. The segmentation version applies Feature Propagation layers to upsample. The PointNet++ networks we use in experiments have approximately 1.4M parameters.  The PyTorch print string of the model with 3D flow output is:

{\footnotesize
\begin{verbatim}
Actor(
  PointNet2_Flow(
    (sa1_module): SAModule(
      (conv): PointNetConv(local_nn=MLP(5, 64, 64, 128), global_nn=None)
    )
    (sa2_module): SAModule(
      (conv): PointNetConv(local_nn=MLP(131, 128, 128, 256), global_nn=None)
    )
    (sa3_module): GlobalSAModule(
      (nn): MLP(259, 256, 512, 1024)
    )
    (fp3_module): FPModule(
      (nn): MLP(1280, 256, 256)
    )
    (fp2_module): FPModule(
      (nn): MLP(384, 256, 128)
    )
    (fp1_module): FPModule(
      (nn): MLP(130, 128, 128, 128)
    )
    (mlp): MLP(128, 128, 128, 3)
  )
)
\end{verbatim}
}
To implement the differentiable, parameter-less Singular Value Decomposition (SVD) layer, we use PyTorch3D~\cite{ravi2020pytorch3d}.

\subsubsection{Implementation of \toolflow Ablations}

Below, we expand upon the description of ablations in Section~\ref{ssec:ablations}. 

\begin{itemize}[noitemsep,nolistsep]
\item \textbf{\toolflow, No Skip Connections}: removes the skip connections in the segmentation PointNet++, which we implement by not invoking a \texttt{torch.cat([x, x\_skip])} command in the feature propagation layers~\cite{PointNet2_2017}, where \texttt{x\_skip} represents features from an earlier layer.
\item \textbf{\toolflow, MSE after SVD}: tests applying an MSE loss on the induced transformation from SVD instead of point matching. The output of the SVD is a transformation, which is equivalently expressed as a 6D translation and rotation action vector $\ba$ which we can directly use with MSE against the ground truth actions. We still use the consistency loss (Eq.~\ref{eq:consistency}) to supervise the internal per-point flow vectors, as that may help regularize the predicted flow.
\item \textbf{\toolflow, Point Matching (PM) Before SVD}: tests using the PM loss (Eq.~\ref{eq:point_match}) before the SVD layer, so the loss does not back-propagate through the SVD layer. Here, the SVD is not differentiable, but is used at test time because for a given input point cloud $\pcl$, the output of the neural network is an $(N' \times 3)$-sized flow array, which must then be converted to an action $\ba$. We do not test this ablation with consistency since this would add a second objective to the internal flow predictions which could confuse the network. 
\item \textbf{\toolflow, No Consistency}: tests removing the consistency loss (and just using Eq.~\ref{eq:point_match}), or equivalently, setting $\lambda=0.0$ in $L_{\rm combo}$. This tests to see whether it is feasible to just use the point matching loss without having a loss that directly supervises the predicted flow vectors. We investigate this ablation further in Table~\ref{tab:exp-consistency}.
\end{itemize}

\newpage
\section{Additional Simulation Experiments and Analysis}\label{app:experiments}

We analyze existing experiments and present new ones to further investigate \toolflow in simulation (see Appendix~\ref{app:physical} for physical experiments). Unless stated otherwise, we use the experimental settings from Appendix~\ref{app:evaluation}. 

In Appendix~\ref{app:why-tool-flow-helps}, we present several theories on when using tool flow may be helpful. We investigate these theories in new experiments where we test on translation-only data for \scooping (Appendix~\ref{app:translation-only}) and where we test on a new task to investigate a ``locality hypothesis'' (Appendix~\ref{app:locality}).
Next, we explore hyperparameters and target scaling in Appendix~\ref{app:hyperparameters} and Appendix~\ref{app:scaling-targets} for the main \toolflow method we present.  Building upon these results, we extend our main set of results in Appendix~\ref{app:main-experimental-results}.
We then extend experiments from the paper in Appendix~\ref{app:additional-noise} and Appendix~\ref{app:tool-points} on noise injection and the number of tool points, respectively. Appendix~\ref{app:num-demos} contains new experiments where we test performance based on the training data size.

\subsection{Why is Tool Flow Helpful?}\label{app:why-tool-flow-helps}

Using tool flow as a representation in \toolflow is helpful for our simulation tasks as compared to ``Direct Vector'' representations which directly regress to a single vector. Why is that the case? Building upon our discussion in Section~\ref{ssec:ablations}, we consider two possible theories:

\begin{itemize}[noitemsep,nolistsep]
    \item Tool flow is helpful because it represents rotations in a format that is easier to learn.
    \item Tool flow is helpful because of locality bias.
\end{itemize}

The first theory is relevant to how there are multiple ways to represent rotations. Prior work has shown that naively regressing onto some rotation representations such as quaternions, axis-angles, and Euler angles is challenging~\cite{Continuity_Rotations_2019,rotations_10D_rss_2020,RPMG}, and hence flow may be a representation that induces easier learning.
See Appendix~\ref{app:translation-only} for experiments to probe this theory \hl{in comparison with axis-angle rotations, and see Appendix}~\ref{app:rpmg-rotations} \hl{for experiments with other rotation representations}.

The second theory may be relevant to the object-centric nature of the tasks. In \scooping the policy must reason about the relationship between the ladle and the ball, and in \pouring it must reason about the relationship between the controlled box and the target box. 
See Appendix~\ref{app:locality} for experiments to investigate this theory.

\subsubsection{Learning from Translation-Only Data}\label{app:translation-only}

\begin{table*}[h]
  \setlength\tabcolsep{6.0pt}
  \centering
  \footnotesize
  \begin{tabular}{lcr}
  \toprule
  Method              & Demo Type & \scooping \\ 
  \midrule
  PCL Direct Vector (MSE) & 3DoF & \textbf{0.817$\pm$0.04} \\  
  \toolflow (No SVD)  & 3DoF & \textbf{0.808$\pm$0.04} \\  
  \toolflow           & 3DoF & \textbf{0.769$\pm$0.18} \\  
  \midrule
  PCL Direct Vector (MSE)$^\dagger$ & 4DoF & 0.544$\pm$0.03 \\
  \toolflow$^\dagger$           & 4DoF & \textbf{1.152$\pm$0.07} \\
  \bottomrule
  \end{tabular}
  \\$^\dagger$Results are directly from Table~\ref{tab:bc-results-stderr-v2}. 
  \caption{
  Results on \scooping for 3DoF translation-only demonstrators and, for comparison purposes, the 4DoF demonstration data which is used in other experiments. For each demonstrator type, we bold the best performance numbers from the methods, along with any with overlapping standard errors.
  }
  \vspace{-5pt}
  \label{tab:translation-only}
\end{table*}

To better understand when tool flow as an action representation is beneficial, we run a smaller-scale experiment on \scooping \hl{4D} where we now use a translation-only demonstrator.\footnote{We use \scooping since a policy can succeed without using rotations; this is not the case with \pouring.} We script this demonstrator to lower the ladle, then to translate it in water to get its bowl under the target ball, then to lift the ladle (ideally with the ball); its success rate is 0.832. \hl{This environment uses the same ladle and starting structure as ScoopBall 4D from the paper, and thus does not use the alternative tool used in ScoopBall 6D.} 

For this variant, in addition to the standard \toolflow method, we consider a ``\toolflow (No SVD)'' variant, which is a segmentation PointNet++ that (instead of an SVD layer) ends with an ``averaging layer'' which averages all the predicted tool flow vectors.
We test this variant because if the demonstration data is translation-only, then the SVD layer in \toolflow is supervised to produce the identity rotation, which could be challenging as it would require that all tool flow vectors have the same direction. 
We supervise \toolflow (No SVD) with the MSE loss and we do not use a consistency loss.
We compare with the \hl{PCL} Direct Vector (MSE) baseline, which is the same as \toolflow (No SVD) except it uses a classification PointNet++ instead of a segmentation PointNet++.
From Table~\ref{tab:translation-only}, we find that with 3DoF demonstration data, the naive vector policy actually slightly outperforms both \toolflow and \toolflow (No SVD), indicating that \toolflow brings the most benefits when considering both translations and rotations. The results, however, are fairly close with overlapping standard errors (which we consider as per our evaluation practices in Appendix~\ref{app:evaluation}), as $0.817 - 0.04 = 0.777$ for Direct Vector on the lower end of the interval, which is less than the value at the positive end of \toolflow's interval: $0.769+0.18 = 0.949$.

\subsubsection{Locality Bias Hypothesis}\label{app:locality}

One hypothesis we have on why \toolflow~is better than the naive Direct Vector baselines is that the dense representation in \toolflow~brings better locality bias, i.e., \toolflow~might reason better about the relationship between the tool and target object (e.g., the ball in \scooping, and the target cup in \pouring) when the tool is near the target object. To test this hypothesis, we design a simple task where the goal is to move the tool, represented as a sphere, to the target, represented as another sphere. The action is the 3D translation of the tool sphere, and the reward is the negative distance between the tool and the target sphere. Since the action is translation only, we average the predicted flow from \toolflow~to get the final translation action without using the SVD layer. For the observation, we randomly sample 100 points on the surface of the tool sphere, and another 100 points on the surface of the target sphere. The expert demonstration is simply moving the tool towards the target sphere. As in the main paper, \toolflow~is trained using the point matching loss and the consistency loss. We vary the initial distance between the tool and the target sphere, and 
if the locality hypothesis is true, then we would expect a larger gap between \toolflow~and the naive vector baseline when the tool is near the target, and a smaller gap between  \toolflow~and the naive vector baseline when the tool is far away from the target.

\begin{figure}
    \centering
    \includegraphics[width=0.70\textwidth]{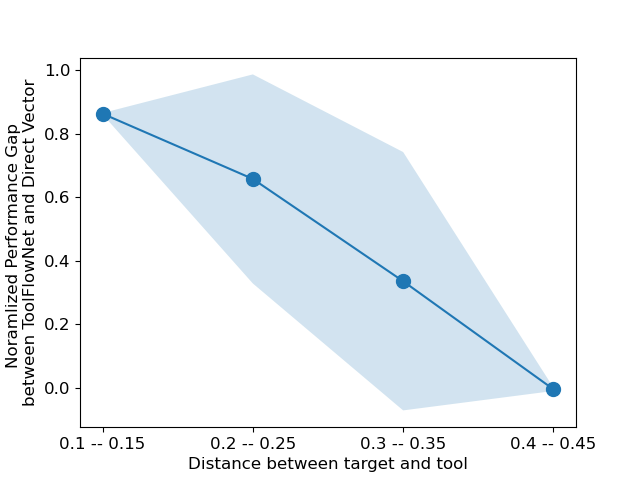}
    \caption{The normalized performance gap between \toolflow~and the Direct Vector baseline, with varying initial distance between the tool and the target sphere. For x-axis ticks, $0.1$ -- $0.15$ means the initial distance is uniformly sampled from the range $[0.1, 0.15]$. }
    \label{fig:locallity}
    \vspace{-0pt}
\end{figure}

As Figure~\ref{fig:locallity} shows, the normalized performance gap between \toolflow~and the baseline (computed as the normalized performance of \toolflow minus the normalized performance of the vector baseline) indeed decreases as the initial distance between the tool and the target increases (results averaged across 5 seeds), which provides support for the locality bias hypothesis. 

\subsection{Consistency Loss Hyperparameter}\label{app:hyperparameters}

\begin{table*}[h]
  \setlength\tabcolsep{6.0pt}
  \centering
  \footnotesize
  \begin{tabular}{lccccc}
  \toprule
  Method    &  $\lambda$ & \scooping 4D & \scooping 6D & \pouring 3D & \pouring 6D \\ 
  \midrule
  \toolflow & 0.0 & 0.861$\pm$0.04 & \hl{0.744$\pm$0.12} & 0.468$\pm$0.09 & \hl{\textbf{0.609$\pm$0.06}} \\
  \toolflow & 0.1 & \textbf{1.152$\pm$0.05} & \hl{\textbf{0.952$\pm$0.02}} & \textbf{0.768$\pm$0.04} & \hl{\textbf{0.667$\pm$0.03}} \\
  \toolflow & 0.5 & 0.987$\pm$0.02 & \hl{\textbf{0.912$\pm$0.04}} & \textbf{0.795$\pm$0.05} & \hl{\textbf{0.658$\pm$0.09}} \\
  \toolflow & 1.0 & 0.873$\pm$0.05 & \hl{\textbf{0.944$\pm$0.03}} & \textbf{0.777$\pm$0.04} & \hl{\textbf{0.628$\pm$0.05}} \\
  \bottomrule
  \end{tabular}
  \caption{
  Performance of \toolflow \hl{on all combinations of the environments and action spaces}. \hl{We vary the} consistency weight $\lambda$ for the consistency loss in Eq.~\ref{eq:consistency}. We bold the best results in each column along with those that have overlapping standard errors. 
  }
  \vspace{-0pt}
  \label{tab:exp-consistency}
\end{table*}
We test different values of the $\lambda$ weight for the consistency loss: ${\lambda \in \{0.0, 0.1, 0.5, 1.0\}}$. Using $\lambda=0$ is the same as the ablation named ``\toolflow, No Consistency'' in Table~\ref{tab:bc-results-stderr-v2}. In Table~\ref{tab:exp-consistency} we report the BC test-time performance (using the standard metric of 5 independent BC runs and taking the best epoch) for both tasks. We find that for \hl{both action spaces of} \scooping, the best value seems to be $\lambda=0.1$ (and leads to slightly outperforming the demonstrator). For \pouring \hl{3D and 6D, the best values are $\lambda=0.5$ and $\lambda=0.1$, respectively} though there are multiple values of $\lambda$ for which performance is similar based on the range of standard errors. \hl{Consequently}, in other experiments (such as the ones we report in Table~\ref{tab:bc-results-stderr-v2}) we use $\lambda=0.1$ for \scooping \hl{4D and 6D}, $\lambda=0.5$ for \pouring \hl{3D, and $\lambda=0.1$ for PourWater 6D} for \toolflow.

\subsection{Scaling Targets During Training}\label{app:scaling-targets}

\begin{table*}[h]
  \setlength\tabcolsep{6.0pt}
  \centering
  \footnotesize
  \begin{tabular}{lccc}
  \toprule
  Success      &  Scale? & \scooping \hl{4D} & \pouring \hl{3D} \\ 
  \midrule
  \toolflow$^\dagger$ & \cmark & 1.152$\pm$0.07 & 0.795$\pm$0.05 \\ 
  Direct Vector (MSE)$^\dagger$ & \cmark & 0.544$\pm$0.03 & 0.530$\pm$0.08 \\
  \toolflow           & \xmark & 0.722$\pm$0.14 & 0.706$\pm$0.02 \\
  Direct Vector (MSE) & \xmark & 0.000$\pm$0.00 & 0.000$\pm$0.00 \\
  \bottomrule
  \end{tabular}
  \\$^\dagger$These numbers are directly from Table~\ref{tab:bc-results-stderr-v2}.
  \caption{
  Success rates of \toolflow and the Direct Vector baseline based on whether targets are scaled, by 250X, or kept at defaults (scale 1) with the latter resulting in per-component values within $\pm$0.004.
  }
  \vspace{-0pt}
  \label{tab:exp-scaling}
\end{table*}

One strategy to improve training of \toolflow is to scale the flow vector targets. In simulation, each time step is a single continuous action which results in extremely small changes in the translation and rotation of the tool pose. (In Section~\ref{app:sim-tasks} we state quantitative numbers.) For our experiments, we scale the flow targets by a factor of 250X to empirically get flow vector magnitudes to be bounded by approximately -1 and 1 in each of the three coordinate dimensions. We do not scale the input point cloud $\pcl$ for the forward pass through the PointNet++ network, but we \emph{do} scale it to ensure correctness in computing the point matching loss in Eq.~\ref{eq:point_match}, since the computation must be done with all values expressing the same units. Intuitively, the scaling acts as a way of converting the units to make the raw values more suitable for training (\eg going from meters to millimeters).

To verify our design choice, we run an experiment where we test \toolflow (with the point matching and consistency loss), with and without scaling. We run 5X \hl{Behavioral Cloning} runs, and report the average (and standard error of the mean) of the best epoch. The results in Table~\ref{tab:exp-scaling} suggest clear benefits to scaling the flow vectors in both tasks.

For consistency, we perform a similar scaling for the baseline, non-flow methods by multiplying their translational magnitudes by 250X. We also perform a similar scaling of the rotations to get their values to be roughly the same order of magnitude as the translation magnitudes. 
Consider the ``Direct Vector'' method, which uses a classification PointNet++ to directly regresses to the action vector. We supervise this with the MSE loss. The results with and without scaling, also in Table~\ref{tab:exp-scaling}, show that without scaling, the training collapses and the performance is zero. Nonetheless, despite strengthening the baseline with scaling of the targets, it remains worse versus \toolflow.

When scaling targets, we ``undo'' the scaling at inference time when performing test rollouts.

\subsection{Main Experimental Results}\label{app:main-experimental-results}

\begin{table*}[t]
  \setlength\tabcolsep{5.0pt}
  \centering
   \scriptsize
    \begin{tabular}{lcccccccc}
    \toprule
    Method & Loss & Dense & \hl{N.} Success & \hl{N.} Success & \hl{N.} Success & \hl{N.} Success & Average \\ 
    & & PN++? & \scooping 4D & \scooping 6D & \pouring 3D & \pouring 6D & \hl{N. Success} \\ 
    \midrule
    \hl{PCL} Direct Vector          & MSE & \xmark & 0.408$\pm$0.01 & 0.640$\pm$0.03 & 0.337$\pm$0.04 & 0.264$\pm$0.01 & 0.412 \\  
    \hl{PCL} Direct Vector          & PM  & \xmark & 0.128$\pm$0.07      & \hl{0.002$\pm$0.00}  & 0.045$\pm$0.02 & \hl{0.042$\pm$0.01}  & \hl{0.055} \\
    \hl{PCL} Dense Transformation   & MSE & \cmark & 0.427$\pm$0.02 & 0.669$\pm$0.06 & 0.372$\pm$0.02 & 0.212$\pm$0.03 & 0.420 \\  
    \hl{PCL} Dense Transformation   & PM  & \cmark & 0.235$\pm$0.03      & \hl{0.158$\pm$0.04}  & 0.316$\pm$0.01 & \hl{0.020$\pm$0.01}  & \hl{0.182} \\
    \hl{D Direct Vector}            & MSE & \xmark & 0.119$\pm$0.03 & 0.744$\pm$0.02 &  0.013$\pm$0.00 &  0.020$\pm$0.00 &  0.224 \\
    \hl{D+S Direct Vector}          & MSE & \xmark & \hl{0.311$\pm$0.04} & \hl{0.804$\pm$0.03} &  \hl{0.656$\pm$0.03} &  \hl{0.231$\pm$0.01} &  \hl{0.500}  \\
    \hl{RGB Direct Vector}          & MSE & \xmark & 0.213$\pm$0.03      & \hl{0.646$\pm$0.01}  & 0.607$\pm$0.03 & \hl{0.216$\pm$0.01}  & \hl{0.420} \\
    \hl{RGB+S Direct Vector}        & MSE & \xmark & \hl{0.326$\pm$0.03} & \hl{\textbf{0.872$\pm$0.02}} & \hl{\textbf{0.734$\pm$0.01}} & \hl{0.179$\pm$0.01} & \hl{0.528}  \\
    \hl{RGBD Direct Vector}         & MSE & \xmark & \hl{0.263$\pm$0.03} & \hl{0.817$\pm$0.02} & \hl{0.662$\pm$0.02} & \hl{0.221$\pm$0.01} & \hl{0.491} \\
    \hl{RGBD+S Direct Vector}       & MSE & \xmark & \hl{0.423$\pm$0.04} & \hl{\textbf{0.883$\pm$0.02}} & \hl{\textbf{0.713$\pm$0.03}} & \hl{0.227$\pm$0.01} & \hl{0.561}  \\
    \midrule
    \toolflow, No Skip Conn.  & PM+C & \cmark & \textbf{0.768$\pm$0.03} & \hl{0.130$\pm$0.02} & 0.000$\pm$0.00 & \hl{0.000$\pm$0.00} & \hl{0.225} \\
    \toolflow, MSE after SVD  & MSE+C & \cmark & 0.011$\pm$0.01          & \hl{0.643$\pm$0.04} & 0.324$\pm$0.01 & \hl{\textbf{0.604$\pm$0.04}} & \hl{0.395} \\
    \toolflow, PM before SVD  & PM  & \cmark & 0.550$\pm$0.04          & \hl{0.708$\pm$0.03} & 0.410$\pm$0.02 & \hl{0.430$\pm$0.02} & \hl{0.525} \\
    \toolflow, No Consistency & PM  & \cmark & 0.585$\pm$0.04          & \hl{0.461$\pm$0.04} & 0.289$\pm$0.11 & \hl{0.375$\pm$0.03} & \hl{0.427} \\
    \midrule
    \textbf{\toolflow (Ours)} & PM+C  & \cmark & \textbf{0.813$\pm$0.02} & \hl{0.799$\pm$0.02} & 0.692$\pm$0.03 & \hl{0.536$\pm$0.01} & \hl{\textbf{0.710}} \\
    \bottomrule
    \end{tabular}
    \caption{
  Results from the same set of experiments reported in Table~\ref{tab:bc-results-stderr-v2}, except we use a different evaluation metric, based on averaging the normalized test-time success rate across all \hl{evaluation (every 25)} epochs, instead of picking the best one epoch. Hence, the raw numbers are lower. See Appendix~\ref{app:main-experimental-results} for further details. 
  \hl{We bold the best numbers in the columns, plus any with overlapping standard errors.} 
  }
  \vspace*{-5pt}
  \label{tab:bc-results-stderr-v3}
\end{table*}

We report the main set of experimental results in Table~\ref{tab:bc-results-stderr-v2} with the evaluation metrics described in Appendix~\ref{app:evaluation}. As there are five independent BC runs, we report standard errors of the mean for each normalized success rate metric.


The results indicate that \toolflow outperforms baselines and ablations, \hl{on average, across all the task and action variants. It has the highest average normalized success rate of 0.892, while the next highest baseline is RGBD+S Direct Vector with an average of 0.753 across the four evaluated tasks and actions.} 

We also observe an intriguing result with the no skip connection ablation of \toolflow, in that it has strong performance on \scooping but never succeeds on \pouring. From inspecting the policy rollouts, we find that without skip connections, the policy cannot perform any rotations. \hl{This occurs because if there are no skip connections, then in the upsampling procedure of segmentation PointNet++ (i.e., the interpolation layers), the same latent vector is copied to every point, so the final predicted flow is the same for every point. When SVD converts the flow to a transformation, this results in a translation-only transformation with no rotation. Upon further analysis, this is due to global pooling layer in the middle of the architecture}~\cite{PointNet2_2017}.

For further analysis, Table~\ref{tab:bc-results-stderr-v3} reports the same experimental runs and settings as Table~\ref{tab:bc-results-stderr-v2}, except with \hl{an alternative} evaluation metric. \hl{To clarify, other than for the current analysis in this subsection, we do \emph{not} use this alternative evaluation metric anywhere else in the paper.} Here, instead of taking the best snapshot among all saved snapshots (each is associated with an epoch, and saved once every 25 epochs), we average the normalized performance across all 20 epochs from 25, 50, and so on, up to 500, and take another average over random seeds, and report that. The advantage of this metric is that it may be more robust to noisy evaluation rollouts as it would average across the full training history. Moreover, it can be useful if one cares more about convergence speed. \hl{From analyzing Tables}~\ref{tab:bc-results-stderr-v2} \hl{and}~\ref{tab:bc-results-stderr-v3}, we find that the results are consistent among both evaluation metrics, with both suggesting that \toolflow outperforms other methods. \hl{From Table}~\ref{tab:bc-results-stderr-v3}, \hl{ToolFlowNet gets the highest average normalized success of 0.710. The next best method is RGBD+S Direct Vector again, with 0.561 average success.}

\hl{For a more complete set of results, we also present additional tables that show the \emph{raw} success rate instead of the normalized success rate. The results with the raw success rate are shown in Table}~\ref{tab:bc-results-RAW-maxepoch} \hl{which corresponds to normalized results in Table}~\ref{tab:bc-results-stderr-v2}, \hl{and Table}~\ref{tab:bc-results-RAW-allepochs}, which corresponds to normalized results in Table~\ref{tab:bc-results-stderr-v3}.

\begin{table*}[t]
  \setlength\tabcolsep{5.0pt}
  \centering
   \scriptsize
    \begin{tabular}{lcccccccc}
    \toprule
    Method & Loss & Dense & R. Success & R. Success & R. Success & R. Success & Average \\ 
    & & PN++? & \scooping 4D & \scooping 6D & \pouring 3D & \pouring 6D & R. Success \\ 
    &  &  & Demo: 0.632 & Demo: 1.000  & Demo: 0.906 & Demo: 0.815 & \\ 
    \midrule
    PCL Direct Vector          & MSE & \xmark & 0.344$\pm$0.02 & 0.848$\pm$0.05 & 0.480$\pm$0.07 & 0.328$\pm$0.03 &  0.500 \\ 
    PCL Direct Vector          & PM  & \xmark & 0.144$\pm$0.08 &  0.048$\pm$0.04 &  0.120$\pm$0.07 & 0.072$\pm$0.03 &  0.096 \\
    PCL Dense Transformation   & MSE & \cmark & 0.328$\pm$0.04 & 0.824$\pm$0.06 & 0.488$\pm$0.04 & 0.280$\pm$0.02 &  0.480 \\ 
    PCL Dense Transformation   & PM  & \cmark & 0.232$\pm$0.04 &  0.360$\pm$0.10  &  0.528$\pm$0.03 &  0.040$\pm$0.02 &  0.290 \\
    D Direct Vector            & MSE & \xmark & 0.120$\pm$0.05 &  \textbf{0.952$\pm$0.02} &  0.032$\pm$0.01 &  0.056$\pm$0.02  &  0.290 \\
    D+S Direct Vector          & MSE & \xmark & 0.464$\pm$0.07 &  \textbf{0.928$\pm$0.03} &  \textbf{0.704$\pm$0.03} &  0.248$\pm$0.02 &  0.586 \\
    RGB Direct Vector          & MSE   & \xmark & 0.224$\pm$0.03 &  0.776$\pm$0.05 &  0.632$\pm$0.02 &  0.264$\pm$0.04  &  0.474  \\
    RGB+S Direct Vector        & MSE & \xmark & 0.424$\pm$0.04 & \textbf{0.944$\pm$0.02}  &  \textbf{0.728$\pm$0.03} &  0.288$\pm$0.03 &  0.596 \\
    RGBD Direct Vector         & MSE   & \xmark & 0.264$\pm$0.06 & 0.920$\pm$0.02 &  0.664$\pm$0.06 &  0.288$\pm$0.02 & 0.534  \\
    RGBD+S Direct Vector       & MSE & \xmark & 0.464$\pm$0.07 &  \textbf{0.968$\pm$0.02} & \textbf{0.752$\pm$0.03} &  0.392$\pm$0.02 &  0.644 \\
    \midrule 
    \toolflow, No Skip Conn.   & PM+C  & \cmark & 0.624$\pm$0.05 &  0.304$\pm$0.06 &  0.000$\pm$0.00 &  0.000$\pm$0.00 & 0.232 \\
    \toolflow, MSE after SVD   & MSE+C & \cmark & 0.056$\pm$0.03 &  0.792$\pm$0.09 &  0.448$\pm$0.02 &  \textbf{0.744$\pm$0.04} & 0.510 \\
    \toolflow, PM before SVD   & PM    & \cmark & 0.496$\pm$0.05 &  0.880$\pm$0.05 &  0.560$\pm$0.04 &  0.552$\pm$0.04 & 0.622 \\
    \toolflow, No Consistency  & PM    & \cmark & 0.544$\pm$0.04 &  0.744$\pm$0.12 &  0.424$\pm$0.09 &  0.496$\pm$0.05 & 0.552 \\
    \midrule
    \textbf{\toolflow (Ours)}  & PM+C  & \cmark & \textbf{0.728$\pm$0.05} & \textbf{0.952$\pm$0.02} &  \textbf{0.720$\pm$0.05} &  0.544$\pm$0.03 & \textbf{0.736}  \\
    \bottomrule
    \end{tabular}
  \caption{
  \hl{These results are the \emph{raw}, un-normalized success rates (R. Success) for the same set of experiments reported in Table}~\ref{tab:bc-results-stderr-v2}, \hl{which normalizes success rates by dividing them by the raw demonstrator performance. The number after ``Demo:'' in the table shows the raw demonstrator success rate. Since ScoopBall 6D has a demonstrator performance of 1.000, the raw values are the same as the normalized values reported in Table}~\ref{tab:bc-results-stderr-v2}, \hl{but the other columns will show different values}.
  }
  \vspace*{-0pt}
  \label{tab:bc-results-RAW-maxepoch}
\end{table*}  

\begin{table*}[t]
  \setlength\tabcolsep{5.0pt}
  \centering
   \scriptsize
    \begin{tabular}{lcccccccc}
    \toprule
    Method & Loss & Dense & R. Success & R. Success & R. Success & R. Success & Average \\ 
    & & PN++? & \scooping 4D & \scooping 6D & \pouring 3D & \pouring 6D & R. Success \\ 
    &  &  & Demo: 0.632 & Demo: 1.000  & Demo: 0.906 & Demo: 0.815 & \\ 
    \midrule
    PCL Direct Vector          & MSE   & \xmark & 0.258$\pm$0.01 & 0.640$\pm$0.03 & 0.305$\pm$0.04 & 0.215$\pm$0.01 & 0.354 \\ 
    PCL Direct Vector          & PM    & \xmark & 0.081$\pm$0.04  &  0.002$\pm$0.00 &  0.041$\pm$0.02 &  0.034$\pm$0.00 &  0.040 \\
    PCL Dense Transformation   & MSE   & \cmark & 0.270$\pm$0.01 & 0.669$\pm$0.06 & 0.337$\pm$0.02 & 0.172$\pm$0.02 & 0.362 \\ 
    PCL Dense Transformation   & PM    & \cmark & 0.148$\pm$0.02 &  0.158$\pm$0.04 &  0.286$\pm$0.01 &  0.016$\pm$0.01 &  0.152  \\
    D Direct Vector            & MSE & \xmark &  0.075$\pm$0.02 &  0.744$\pm$0.02 &  0.012$\pm$0.00 &  0.016$\pm$0.00 &  0.212  \\
    D+S Direct Vector          & MSE   & \xmark & 0.196$\pm$0.03 &  0.804$\pm$0.03 &  0.594$\pm$0.03 &  0.188$\pm$0.01 &  0.446 \\
    RGB Direct Vector          & MSE   & \xmark & 0.134$\pm$0.02 &  0.646$\pm$0.01 &  0.550$\pm$0.02 &  0.176$\pm$0.01 &  0.377 \\
    RGB+S Direct Vector        & MSE & \xmark & 0.206$\pm$0.02 &  \textbf{0.872$\pm$0.02} & \textbf{0.665$\pm$0.01} &  0.146$\pm$0.01 &  0.472 \\
    RGBD Direct Vector         & MSE   & \xmark & 0.166$\pm$0.02 &  0.817$\pm$0.02 &  0.600$\pm$0.02 &  0.180$\pm$0.01 &  0.441 \\
    RGBD+S Direct Vector       & MSE & \xmark & 0.267$\pm$0.03 &  \textbf{0.883$\pm$0.02} &  \textbf{0.646$\pm$0.03} &  0.185$\pm$0.01 &  0.495 \\
    \midrule 
    \toolflow, No Skip Conn.   & PM+C  & \cmark & \textbf{0.485$\pm$0.02} &  0.130$\pm$0.02 &  0.000$\pm$0.00 &  0.000$\pm$0.00 &  0.154 \\
    \toolflow, MSE after SVD   & MSE+C & \cmark & 0.007$\pm$0.00 &  0.643$\pm$0.04 &  0.293$\pm$0.01 &  \textbf{0.492$\pm$0.03} &  0.359  \\
    \toolflow, PM before SVD   & PM    & \cmark & 0.348$\pm$0.01 &  0.708$\pm$0.03 &  0.372$\pm$0.02 &  0.351$\pm$0.02 &  0.444 \\
    \toolflow, No Consistency  & PM    & \cmark & 0.370$\pm$0.02 &  0.461$\pm$0.04 &  0.262$\pm$0.10 &  0.306$\pm$0.03 &  0.349 \\
    \midrule
    \textbf{\toolflow (Ours)}  & PM+C  & \cmark & \textbf{0.514$\pm$0.01} &  0.799$\pm$0.02 &  0.627$\pm$0.01 & 0.437$\pm$0.01 &  \textbf{0.594}  \\
    \bottomrule
    \end{tabular}
  \caption{
  \hl{The raw, un-normalized results from Table}~\ref{tab:bc-results-stderr-v3} \hl{which reports normalized success rates computed by taking the average performance over all evaluation epochs (instead of taking the maximum as in Tables}~\ref{tab:bc-results-stderr-v2} \hl{and}~\ref{tab:bc-results-RAW-maxepoch}). \hl{As with Table}~\ref{tab:bc-results-RAW-maxepoch}, \hl{we repeat the demonstrator raw performance after ``Demo:'' in the relevant columns.}
  }
  \vspace*{-0pt}
  \label{tab:bc-results-RAW-allepochs}
\end{table*}

\subsection{Deep Reinforcement Learning Baseline}\label{app:deep-rl-baseline}

\begin{figure}[t]
  \centering
  \includegraphics[width=1.00\textwidth]{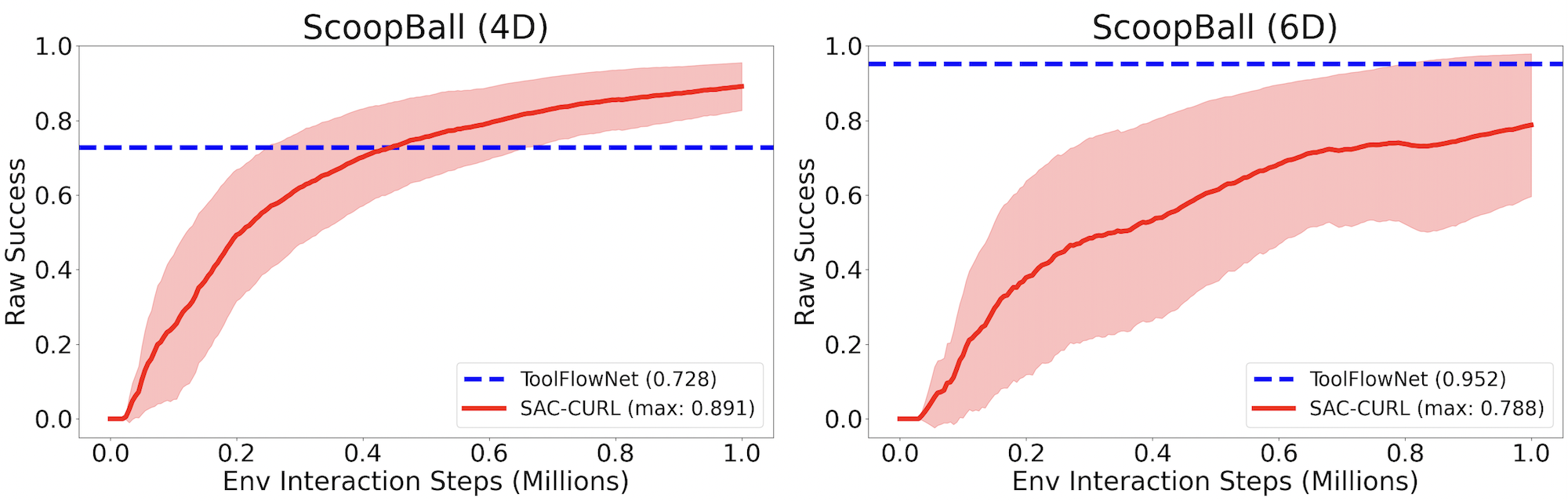}
  \caption{
  \hl{Performance of SAC-CURL on ScoopBall with the two action space variants we test in this paper (4D and 6D). We show the raw (not normalized) test-time success rate, and the curve is smoothed and averaged over 3 random seeds. For comparison, we overlay the performance of ToolFlowNet. The legend contains the performance of ToolFlowNet and the maximum performance along the SAC-CURL performance curve.}
  }
  \vspace{-0pt}
  \label{fig:scoop-ball-RL}
\end{figure}

\begin{figure}[t]
  \centering
  \includegraphics[width=1.00\textwidth]{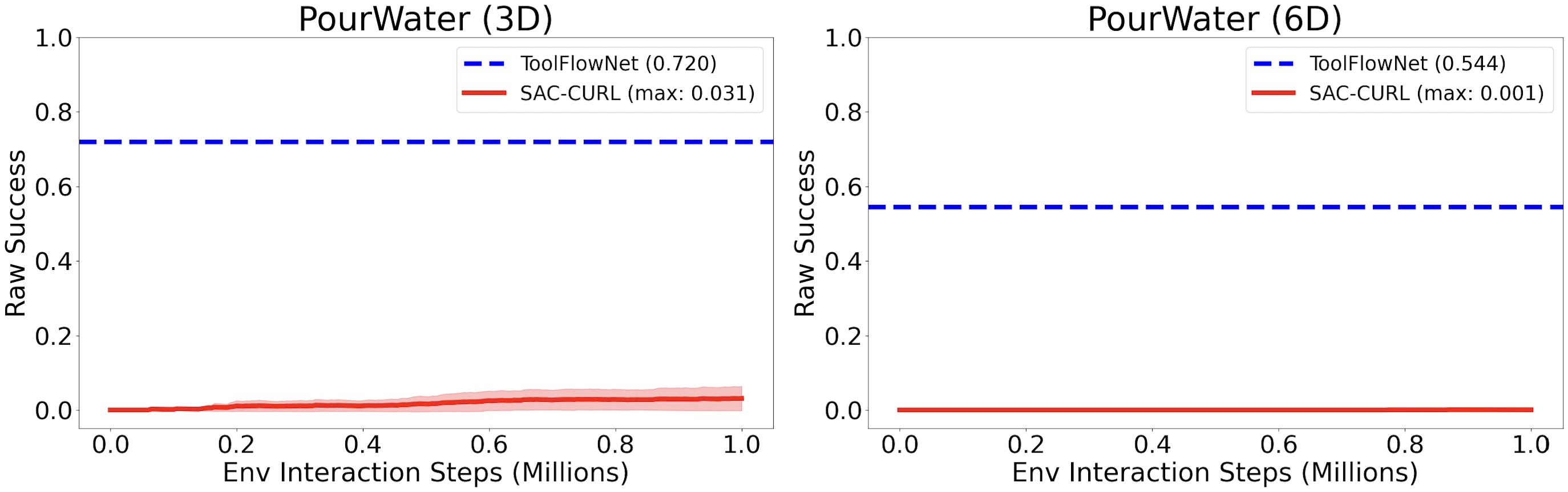}
  \caption{
  \hl{Performance of SAC-CURL on PourWater with the two action space variants we test in this paper (3D and 6D). The plot is formatted in a similar manner as in Figure}~\ref{fig:scoop-ball-RL}.
  }
  \vspace{-0pt}
  \label{fig:pouring-RL}
\end{figure}

\hl{To get a rough sense of how RL compares against IL, we try SAC-CURL}~\cite{CURL} \hl{from the open-source SoftAgent repository\footnote{\url{https://github.com/Xingyu-Lin/softagent}} on the PourWater and ScoopBall environments using RGB image inputs. For both environments, and for both action variants for each environment, we train SAC-CURL for 1 million training steps (i.e., environment interaction steps) and perform 10 test-time evaluation steps every 5000 steps. We run 3 random seeds for each experiment setting. We use dense rewards for both environments. ScoopBall's dense reward is the relative height of the ball, and PourWater's dense reward is the fraction of water particles inside the target.}

\hl{Figures}~\ref{fig:scoop-ball-RL} and~\ref{fig:pouring-RL} \hl{show the SAC-CURL performance curves for ScoopBall and PourWater, respectively. We plot the performance curve for SAC-CURL and smooth it using an exponentially weighted averaging. We also take the performance of the ToolFlowNet policy (using \emph{raw} success, from Table}~\ref{tab:bc-results-RAW-maxepoch}) \hl{and plot its performance in the figures with horizontal dashed lines.}

\hl{On ScoopBall 4D, SAC-CURL obtains a maximum success rate of 0.891 after 1 million training steps. While this is an impressive raw performance, it required over 400,000 steps of environment interaction before surpassing the performance of the ToolFlowNet policy. For ScoopBall 4D, ToolFlowNet learned from 100 demonstration episodes of 100 time steps each, resulting in a total of just 10,000 (offline) state-action pairs. The SAC-CURL policy learns to avoid scooping the water, since accumulating water in the ladle causes unstable physics in that water tends to push the ball out of the ladle's control. This may explains the lower success rate of ToolFlowNet compared to SAC-CURL, because ToolFlowNet was imitating a demonstrator which scooped water.}

\hl{For ScoopBall 6D, ToolFlowNet achieves a higher success rate of 0.952 because it imitates a much more reliable demonstrator and uses a ladle which allows water to pass through it, which addresses some physics instability. The 0.952 value is higher than the final average achieved by SAC-CURL (0.788) even after 1 million environment steps.}

\hl{The results for PourWater for both action variants show an even more pronounced benefit for imitation learning using ToolFlowNet over SAC-CURL. Even after 1 million environment interaction steps, SAC-CURL gets \emph{close to zero} binary success rate for both variants of PourWater, whereas ToolFlowNet is significantly more reliable with raw success rates of 0.720 and 0.544 for 3 DoF and 6 DoF action spaces, respectively.
}

\hl{As shown on the project website\footnote{\url{https://tinyurl.com/toolflownet}}, the policies learned from SAC-CURL tend to qualitatively look jerkier and more unstable compared to policies from imitation learning. Overall, these results may provide evidence for the benefits of imitation learning in these environments over reinforcement learning. An interesting future direction to explore for these tasks would be to combine imitation learning with reinforcement learning}~\cite{DMfD_2022,DQfD,Overcoming_Exploration_2018}.

\subsection{State-Based Policy Baseline}\label{app:state-based-policy-baseline}

\begin{table*}[h]
  \setlength\tabcolsep{5.0pt}
  \centering
  \footnotesize
  \begin{tabular}{lccccc}
  \toprule
  Method &  ScoopBall 4D & ScoopBall 6D & PourWater 3D & PourWater 6D & Average \\ 
  \midrule
  State (G.T.) Direct Vector & \textbf{1.152$\pm$0.04} &  0.336$\pm$0.06 & \textbf{0.768$\pm$0.02} & \textbf{ 0.785$\pm$0.07} & 0.760 \\
  State (Learned) Direct Vector & 0.835$\pm$0.12 & 0.824$\pm$0.06 & 0.433$\pm$0.07 & 0.226$\pm$0.03 & 0.579 \\
  \toolflow$^\dagger$     & \textbf{1.152$\pm$0.07} & \textbf{0.952$\pm$0.02}  & \textbf{0.795$\pm$0.05} & 0.667$\pm$0.03 & \textbf{0.892} \\
  \bottomrule
  \end{tabular}
  \\$^\dagger$These results are directly from Table~\ref{tab:bc-results-stderr-v2}.
  \caption{
  \hl{Normalized success rates on the four task and action space combinations explored in the paper. We compare ToolFlowNet with state-based policies; see Section}~\ref{app:state-based-policy-baseline} \hl{for more details.}
  }
  \vspace{-0pt}
  \label{tab:state-policy}
\end{table*}

\hl{As another set of baselines, we consider state-based policies which assume access to ground-truth tool and object poses. For ScoopBall, the state input is a concatenated vector of the 7D ladle pose (position and quaternion) and the 3D center of the ball, resulting in a 10D state vector. For PourWater, the state input is a concatenated vector of the state of the controlled box and the target box. Each box has 11 values in its state: its 3D center position, its 4D quaternion, its 3D dimensions (width, length, and height), and a 1D scalar representing the fraction of water particles in it. With two boxes, the state vector is thus 22D.}

\hl{We train two variants of state-based methods, called \textbf{State (G.T.) Direct Vector} and \textbf{State (Learned) Direct Vector}, both trained with MSE on the action vectors. For State (G.T.) Direct Vector, we directly use access to the ground truth poses and pass that state information as input to an MLP policy network.  The MLP policy network consists of a fully connected network with two layers of 256 nodes each with ReLU activations, producing a single 6D action vector output.}

\hl{For State (Learned) Direct Vector, we first train a neural network policy which processes segmented point clouds as input and predicts the state information. (The segmented point clouds are the same type of inputs that we provide to ToolFlowNet.)  The neural network policy is a PointNet++ built on the standard ``classification'' architecture for PointNet++.  Then, we fix this network and, in a second training stage, use the output from this network as input to an MLP, which is trained to predict the actions. This MLP has the same architecture as in the State (G.T.) Direct Vector baseline. To clarify, even the ``State (Learned) Direct Vector" baseline requires access to the ground-truth pose of objects in the environment during training in order to train the state estimators, whereas ToolFlowNet does not require access to such ground-truth state information.}

\hl{In Table}~\ref{tab:state-policy}, \hl{we report the normalized success rates of the state-based policy baselines. 
The results suggest that State (G.T.) Direct Vector performs well. It attains similar performance as ToolFlowNet (in that standard error ranges overlap) on ScoopBall 4D and PourWater 3D, outperforms it on PourWater 6D, and performs much worse on ScoopBall 6D, though on average, ToolFlowNet performs slightly better (0.892 versus 0.760). For State (Learned) Direct Vector, performance is worse compared to ToolFlowNet on all experiments, and it only outperforms State (G.T.) Direct Vector on 6D pouring}.

\hl{While State (G.T.) Direct Vector policies are able to achieve similar performance as ToolFlowNet, they assume access to ground-truth tool and object poses (and for PourWater, the fraction of water particles in the boxes). While knowledge of object poses has been be used in prior work for learning 6D pose transformations}~\cite{Orienting_Riemannian_2021,OMA_2011,StableVectorFields_2021,Orientation_Learning_Adaptation_2020}, \hl{ToolFlowNet does \emph{not} require access to such information.} 

\subsection{ToolFlowNet with Non-Segmented Point Clouds}\label{app:segless-tfn}

\begin{figure}[t]
  \centering
  \includegraphics[width=1.00\textwidth]{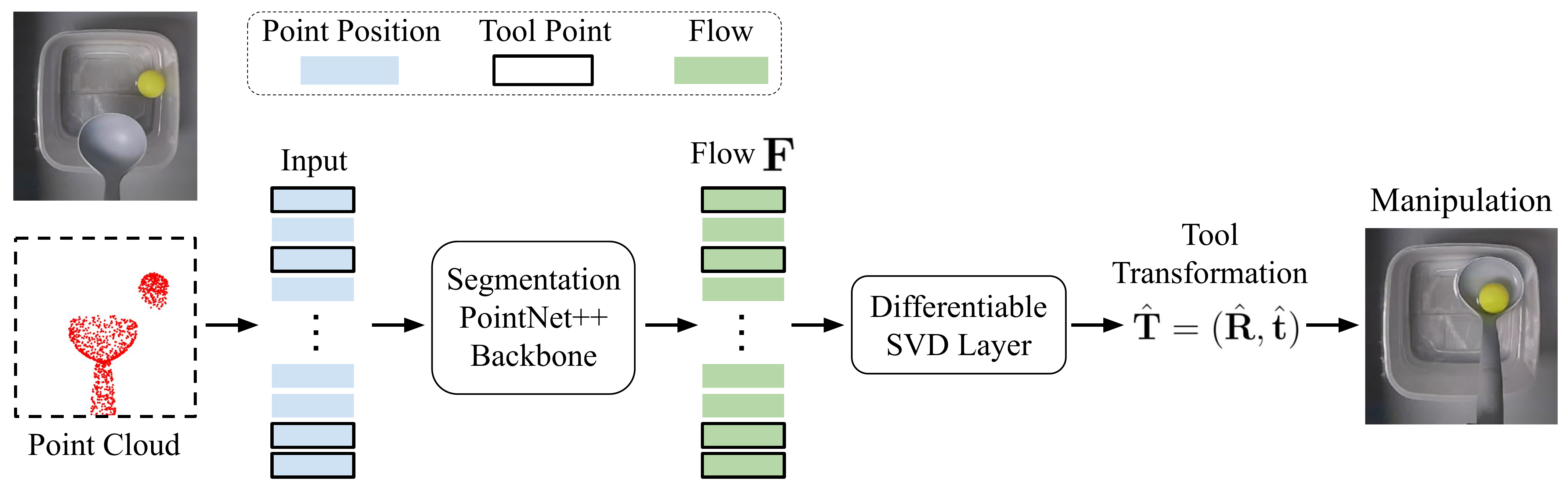}
  \caption{
  \hl{ToolFlowNet but with \emph{non-segmented} point clouds as input. The network only takes in the 3D position coordinates of the input point cloud and uses all points during the SVD layer. We color the input point cloud only for visualization, and outline the tool points in bold to emphasize that both tool and non-tool points are provided to the SVD. See Section}~\ref{app:segless-tfn} \hl{for further details, and Figure}~\ref{fig:system} \hl{for a reference comparison with the standard ToolFlowNet architecture.}
  }
  \vspace{0pt}
  \label{fig:system-segless}
\end{figure}

\begin{table*}[h]
  \setlength\tabcolsep{5.0pt}
  \centering
  \footnotesize
  \begin{tabular}{lccccc}
  \toprule
  Method &  ScoopBall 4D & ScoopBall 6D & PourWater 3D & PourWater 6D & Average \\ 
  \midrule
  \toolflow, Non-Segm  & 0.987$\pm$0.06 &  \textbf{0.928$\pm$0.01} & 0.371$\pm$0.03 & \textbf{0.579$\pm$0.08} & 0.716 \\
  \toolflow$^\dagger$     & \textbf{1.152$\pm$0.07} & \textbf{0.952$\pm$0.02}  & \textbf{0.795$\pm$0.05} & \textbf{0.667$\pm$0.03} & \textbf{0.892} \\
  \bottomrule
  \end{tabular}
  \\$^\dagger$These results are directly from Table~\ref{tab:bc-results-stderr-v2}.
  \caption{
  \hl{Normalized success rates of ToolFlowNet performance without segmented point cloud inputs (``Non-Segm'' in the table) and comparing it with the standard input we use for ToolFlowNet. See Section}~\ref{app:segless-tfn} \hl{for more details.}
  }
  \vspace{-0pt}
  \label{tab:segless-results}
\end{table*}

\hl{We investigate whether we can alleviate a key assumption we make for ToolFlowNet: that we require segmented point cloud observations as input. To modify ToolFlowNet so that it does not use segmentation information, we remove the per-point one-hot classification vector. Thus, the input point cloud consists only of the positions of each point, and has dimension $N\times 3$. Then, in the forward pass, the SVD layer uses the predicted flow from all points (both tool and non-tool).} 
See Figure~\ref{fig:system-segless} \hl{for a visualization of this method}.
\hl{For supervision, we form the per-point, ground-truth flow labels by using a similar method as in the segmented point cloud version (see Section}~\ref{ssec:IL_losses}). \hl{As before, we apply the intended action from the demonstrator to transform points, except we do this to all points, not just the tool points.}

Table~\ref{tab:segless-results} \hl{compares ToolFlowNet results with non-segmented point cloud inputs versus the standard point cloud inputs, under the same experimental conditions and metrics as in Table}~\ref{tab:bc-results-stderr-v2}. 
\hl{We find that, on average, performance without per-point segmentation information is worse, as expected. Nonetheless, in ScoopBall 6D and PourWater 6D, the results with and without segmentation information have overlapping standard errors. Furthermore, the average value of 0.716 for ToolFlowNet without segmentation exceeds the average value for all of the baselines reported in Table}~\ref{tab:bc-results-stderr-v2} \hl{with the exception of RGBD+S Direct Vector, which has an average normalized success of 0.753. This suggests that even without segmentation information, ToolFlowNet can still be effective for imitation learning from point clouds.}

\hl{In addition to these results, we explore another method for learning from segmentation-less point cloud inputs. The SVD layer can utilize learnable per-point weights which indicate how much value to weigh each point during the SVD forward pass. In the standard ToolFlowNet formulation, the weights are not learned and fixed to be 1 for all tool points (thus weighing each tool point equally) and 0 for non-tool points. We adjust this to use learnable weights during ToolFlowNet training, as we hypothesize that there might be enough supervision to learn weights with higher values for tool points and lower values for non-tool points. We implement this by having the forward pass through the segmentation PointNet++ architecture produce four output values, three for the standard flow predictions and one extra value for the per-point weights. We then pass these raw weights through a sigmoid layer, and then through a normalization layer before passing it to the SVD layer. However, the results for this method were worse than the approach presented earlier of assuming that the SVD layer uses all tool and non-tool points, each with equal weight.}

\subsection{Noisy Point Clouds}\label{app:additional-noise}

\begin{figure}[t]
  \centering
  \includegraphics[width=0.75\textwidth]{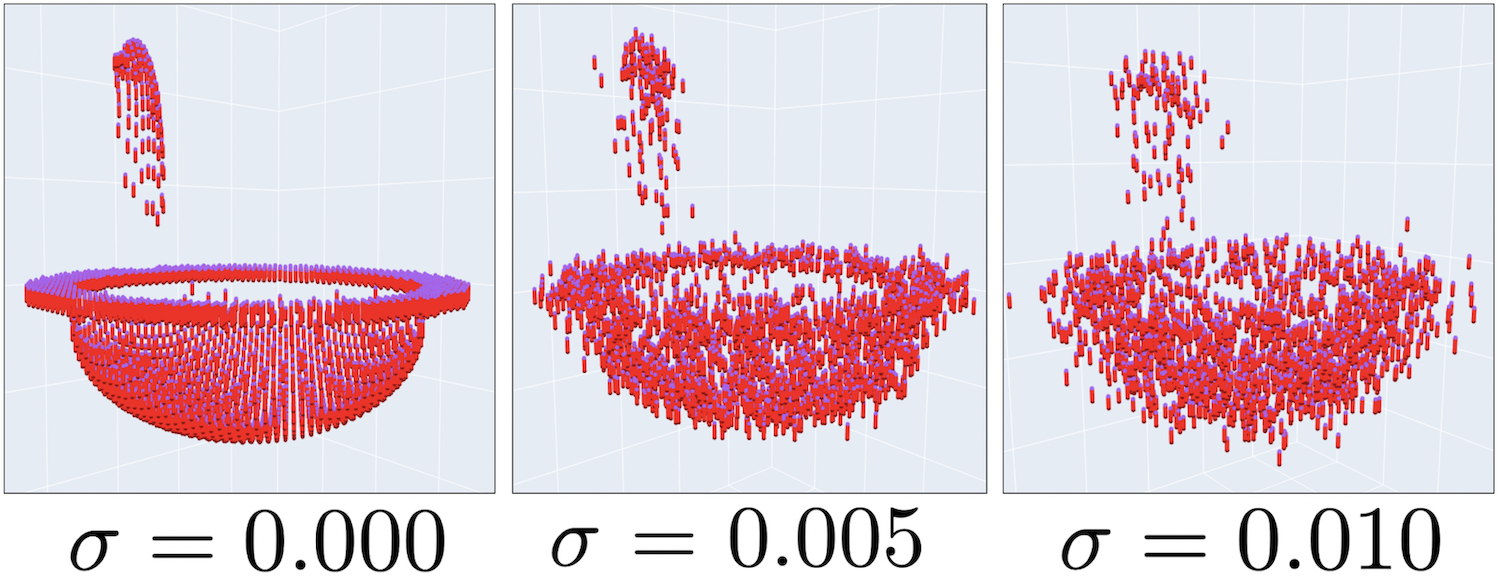}
  \caption{
  Examples of point clouds (blue points) and the corresponding flow (red vectors) visualizations for the tool (\ie the ladle) for \scooping based on different noise injection levels. We test with $\sigma \in \{0.000, 0.005, 0.010\}$ as described in Appendix~\ref{app:additional-noise}. The samples above are from the training data, where the demonstrator happened to perform translation-only actions, so the flow vectors all point downwards and with the same magnitude.
  }
  \vspace{-0pt}
  \label{fig:additional-noise}
\end{figure}

\begin{table*}[h]
  \setlength\tabcolsep{6.0pt}
  \centering
  \footnotesize
  \begin{tabular}{lcccc}
  \toprule
  Method        &  StDev $\sigma$ & \scooping \hl{4D} & \pouring \hl{3D} \\ 
  \midrule
  \toolflow           & 0.010 & 0.785$\pm$0.09 & 0.521$\pm$0.08 \\
  \toolflow           & 0.005 & \textbf{1.013$\pm$0.12} & 0.503$\pm$0.06 \\
  \toolflow$^\dagger$ & 0.000 & \textbf{1.152$\pm$0.07} & \textbf{0.795$\pm$0.05} \\
  \bottomrule
  \end{tabular}
  \\$^\dagger$These noise-free results are directly from Table~\ref{tab:bc-results-stderr-v2}.
  \caption{
  Experiments with injecting Gaussian noise into point clouds. We bold the best results in each task's column and those runs with overlapping standard errors.
  }
  \vspace{-0pt}
  \label{tab:exp-gauss-noise}
\end{table*}

\hl{We next study Behavioral Cloning using ToolFlowNet when we alter the nature of the point clouds. To explore the potential for transfer to real settings with noisier sensor readings, we inject independent, identically distributed Gaussian noise to each point's 3D position in all training and testing data point clouds.} 
See Table~\ref{tab:exp-gauss-noise} for results with testing $\sigma \in \{0.000, 0.005, 0.010\}$ with two tasks; the no-noise case of $\sigma=0.000$ is the default setting for other experiments in this paper. 
We only inject noise in the point cloud positions (for the tool and other items), and we do not perturb the demonstrator's tool flow vectors. The noise injection happens once to each point cloud in the training data and is fixed; this is different from adding noise each time a training data is sampled, which is a type of data augmentation. At test time, we apply a similar level of noise injection to each (new) point cloud observation.

It may be more meaningful to interpret $\sigma$ values by comparing them with the size of the tool, since in simulation we can make the tool of arbitrary size. The value of 0.005 units in simulation is 2.7\% of the radius of the ladle’s bowl’s for \scooping and 2.1\% of the average box length in \pouring, where for the latter, we refer to the box that is the tool (the target box has similar dimensions). This is the average box length; in \pouring the box sizes are randomized, whereas in \scooping the ladle is of a fixed size. 
For visualizations of different noise injections, see Figure~\ref{fig:additional-noise}. This shows both point clouds (in blue) and the ground-truth flow vectors (in red) from the \scooping demonstrations.

The results suggest that \toolflow may be robust to some levels of noise. In particular, for \scooping \hl{4D}, using $\sigma=0.005$ means the best performance is 1.013 and nearly matches the 1.152 performance of the method in the noise-free case (both slightly outperform the demonstrator). As expected, in general with increasing noise, performance deteriorates, though interestingly, in \pouring \hl{3D}, using $\sigma=0.010$ is slightly better than $\sigma=0.005$.

\subsection{Fewer Tool Points}\label{app:tool-points}

\begin{table*}[h]
  \setlength\tabcolsep{6.0pt}
  \centering
  \footnotesize
  \begin{tabular}{lrc}
  \toprule
  Method and Task     &  \#tool & Performance \\ 
  \midrule
  \toolflow, \scooping \hl{4D} &    10 & \textbf{1.063$\pm$0.09}  \\
  \toolflow, \scooping \hl{4D}$^\dagger$ & 1284$^\mathsection$ & \textbf{1.152$\pm$0.07}  \\
  \midrule 
  \toolflow, \pouring \hl{3D}  &    10   & \textbf{0.883$\pm$0.02} \\
  \toolflow, \pouring \hl{3D}$^\dagger$  & 633$^\mathsection$ & 0.795$\pm$0.05 \\
  \bottomrule
  \end{tabular}
  \\$^\dagger$Results are directly from Table~\ref{tab:bc-results-stderr-v2}. 
  \\$^\mathsection$Represents the average number of tool points in a point cloud.
  \caption{
  Performance of \toolflow based on using either a subset of 10 tool points, or the standard number of observable tool points. 
  }
  \vspace{-0pt}
  \label{tab:exp-tool-points}
\end{table*}

\hl{In these experiments, we investigate the performance of ToolFlowNet while using} different numbers of tool points in the point cloud. 
We use 10 points for \pouring \hl{3D}, based on 10 fixed keypoints located on the box. For \scooping \hl{4D}, we similarly use 10 keypoints located on the ladle. These form the 10 tool points in the segmented point cloud. In contrast, for experiments from Table~\ref{tab:bc-results-stderr-v2}, \hl{the ScoopBall and PourWater data have an average of 1284 and 633 tool points, respectively, per observation.} 

See Table~\ref{tab:exp-tool-points} for results. Interestingly, using just 10 tool points seems to be sufficient for \toolflow to imitate the demonstration data. Indeed, the version with \pouring even outperforms the one with the usual amount of tool points with 0.883 normalized performance versus 0.795. 

This result should be interpreted with some nuance. First, we assume these 10 points are always available in the point cloud $\pcl$, even if they are occluded, which is in contrast to the standard experimental setup in this work where we use the observable point cloud, and hence, parts of the tool can be occluded. For example, in \pouring, the tool box frequently occludes itself, and when it gets close to the target box, the target box can also occlude parts of the tool box. Second, in order to get 5 complete Behavioral Cloning runs as per our evaluation metric in Appendix~\ref{app:evaluation}, we had to run the 10 tool point case for \pouring 8 times. Of the 8 initial runs, 3 crashed due to an ill-conditioned matrix input to Singular Value Decomposition (SVD). This may suggest that extra \hl{tool} points can add robustness to the SVD procedure and thus to \toolflow, as we have never encountered this error in other experiments.

\subsection{Number of Training Demonstrations}
\label{app:num-demos}

\begin{table*}[h]
  \setlength\tabcolsep{6.0pt}
  \centering
  \footnotesize
  \begin{tabular}{lrcc}
  \toprule
  Method       &  \# Demos & \scooping \hl{4D} & \pouring \hl{3D} \\ 
  \midrule
  \toolflow    &  10 & 0.620$\pm$0.22 & 0.256$\pm$0.07 \\
  \toolflow    &  50 & \textbf{0.975$\pm$0.11} & 0.477$\pm$0.05 \\
  \toolflow$^\dagger$ & 100 & \textbf{1.152$\pm$0.07} & \textbf{0.795$\pm$0.05} \\
  \bottomrule
  \end{tabular}
  \\$^\dagger$Results are directly from Table~\ref{tab:bc-results-stderr-v2}. 
  \caption{
  Performance of \toolflow as a function of the number of training data demonstrations. 
  }
  \vspace{-0pt}
  \label{tab:exp-demo-data}
\end{table*}

We standardize on 100 training demonstrations for simulation experiments for all tasks and demonstrations with the exception of the 6DoF ScoopBall task where we use 25 demonstrations.  This is mainly due to the different tool which makes the task easier for policy learning. Here, we investigate the performance of \toolflow as a function of the number of training data demonstrations for \scooping 4D and \pouring 3D.
See Table~\ref{tab:exp-demo-data} for the results, which indicate that while performance decreases with fewer demonstrations (as expected), \toolflow can still be more sample efficient than alternative methods. In particular, for \scooping 4D, using \emph{just 10 demonstrations} leads to a normalized success rate of 0.620, which outperforms other baselines from Table~\ref{tab:bc-results-stderr-v2}.

\subsection{Baselines: Local vs Global Coordinates for Axis-Angle Rotations}
\label{app:intrinsic-extrinsic}

\begin{table*}[h]
  \setlength\tabcolsep{5.0pt}
  \centering
  \scriptsize
  \begin{tabular}{lcccccc}
  \toprule
  Method &  Frame & ScoopBall 4D & ScoopBall 6D & PourWater 3D & PourWater 6D & Average \\ 
  \midrule
  PCL Direct Vector (MSE)$^\dagger$        & Local  & 0.544$\pm$0.03 & 0.848$\pm$0.05 & 0.530$\pm$0.08 & 0.402$\pm$0.04 & 0.581 \\
  PCL Direct Vector (MSE)        & Global & 0.519$\pm$0.08 & 0.824$\pm$0.04 & 0.459$\pm$0.04 & 0.167$\pm$0.08 & 0.492 \\
  \midrule
  PCL Dense Transformation (MSE)$^\dagger$ & Local  & 0.519$\pm$0.07 & 0.824$\pm$0.06 & 0.539$\pm$0.05 & 0.344$\pm$0.03 & 0.556 \\
  PCL Dense Transformation (MSE) & Global & 0.646$\pm$0.09 & 0.824$\pm$0.03 & 0.494$\pm$0.05 & 0.216$\pm$0.02 & 0.545 \\
  \midrule
  \toolflow$^\dagger$            & N/A    & \textbf{1.152$\pm$0.07} & \textbf{0.952$\pm$0.02}  & \textbf{0.795$\pm$0.05} & \textbf{0.667$\pm$0.03} & \textbf{0.892} \\
  \bottomrule
  \end{tabular}
  \\$^\dagger$These results are directly from Table~\ref{tab:bc-results-stderr-v2}.
  \caption{
  \hl{Normalized success rates on the four task and action space combinations explored in the paper. We compare baseline methods of PCL Direct Vector and PCL Dense Transformation based on whether the target rotation (in axis-angle format) is expressed in local versus global coordinates. See Section}~\ref{app:intrinsic-extrinsic} \hl{for details.}
  }
  \vspace{-0pt}
  \label{tab:local-global}
\end{table*}

\hl{For the baseline methods of PCL Direct Vector (MSE) and PCL Dense Transformation (MSE), we supervise the models with a 6D target vector, where 3 of the dimensions are for the 3D axis-angle rotation representation. The axis-angle is represented with respect to the local tool frame, centered at the ladle tip (for ScoopBall) or the bottom center part of the box (for PourWater).}

\hl{Concurrent work which studies learning from point clouds has shown how the choice of coordinate frame for the points matter}~\cite{Frame_Mining_2022}. \hl{Motivated by this, we explore whether the baseline methods will improve when we adjust the frame for the axis-angle values, testing \emph{global} axis-angle values with respect to the world frame. We only test with the MSE loss, and do not test the Point Matching loss, as the results from Table}~\ref{tab:bc-results-stderr-v2} \hl{showed that using the MSE loss for the baselines resulted in significantly better success rates.}

\hl{We show the results in Table}~\ref{tab:local-global}, \hl{which also compares against ToolFlowNet. Overall, we find that the choice of coordinate frame for expressing the rotation does not make a significant difference in our tasks. There is a slight boost towards using rotations expressed with respect to the local tool frame, but both baselines remain worse compared to ToolFlowNet.}

\subsection{Baselines: 4D, 6D, 9D, and 10D Rotation Representations}
\label{app:rpmg-rotations}

\begin{table*}[h]
  \setlength\tabcolsep{5.0pt}
  \centering
  \scriptsize
  \begin{tabular}{lcccccc}
  \toprule
  Method &  Rotation & ScoopBall 4D & ScoopBall 6D & PourWater 3D & PourWater 6D & Average \\ 
  \midrule
  PCL Direct Vector (MSE)$^\dagger$ &  3D & 0.544$\pm$0.03 & 0.848$\pm$0.05 & 0.530$\pm$0.08 & 0.402$\pm$0.04 & 0.581 \\
  \midrule
  PCL Direct Vector (MSE)           &  4D & 0.203$\pm$0.05 & 0.280$\pm$0.16 & 0.177$\pm$0.16 & 0.059$\pm$0.05 & 0.180 \\
  PCL Direct Vector (MSE)           &  6D & 0.304$\pm$0.03 & 0.576$\pm$0.14 & 0.212$\pm$0.19 & 0.059 $\pm$0.04 & 0.288 \\
  PCL Direct Vector (MSE)           &  9D & 0.405$\pm$0.11 & 0.320$\pm$0.12 & 0.132$\pm$0.12 & 0.147$\pm$0.09 & 0.251 \\
  PCL Direct Vector (MSE)           & 10D & 0.215$\pm$0.09 & 0.176$\pm$0.16 & 0.079$\pm$0.07 & 0.079$\pm$0.07 & 0.137 \\
  \midrule
  \toolflow$^\dagger$            & N/A    & \textbf{1.152$\pm$0.07} & \textbf{0.952$\pm$0.02}  & \textbf{0.795$\pm$0.05} & \textbf{0.667$\pm$0.03} & \textbf{0.892} \\
  \bottomrule
  \end{tabular}
  \\$^\dagger$These results are directly from Table~\ref{tab:bc-results-stderr-v2}.
  \caption{
  \hl{Experiments comparing normalized test-time success rates of PCL Direct Vector (MSE) with different rotation representations. The 3D rotation represents local axis-angle which we used for results in Table}~\ref{tab:bc-results-stderr-v2}. See Section~\ref{app:rpmg-rotations} for details.
  }
  \vspace{-0pt}
  \label{tab:rpmg-rotations}
\end{table*}

\hl{Prior work}~\cite{Continuity_Rotations_2019,rotations_10D_rss_2020,RPMG} \hl{has demonstrated that regressing to rotations using deep neural networks is challenging with 3D rotation representations such as axis-angle, which we use as our default (non-flow based) rotation representation. We thus perform experiments to check whether using alternative rotation representations can improve performance of the PCL Direct Vector MSE baseline. We test using 4D rotations (quaternions), 6D rotations}~\cite{Continuity_Rotations_2019}, \hl{9D rotations (rotation matrices)}~\cite{levinson20neurips}, \hl{and 10D rotations}~\cite{rotations_10D_rss_2020}.

\begin{figure}[t]
  \centering
  \includegraphics[width=1.00\textwidth]{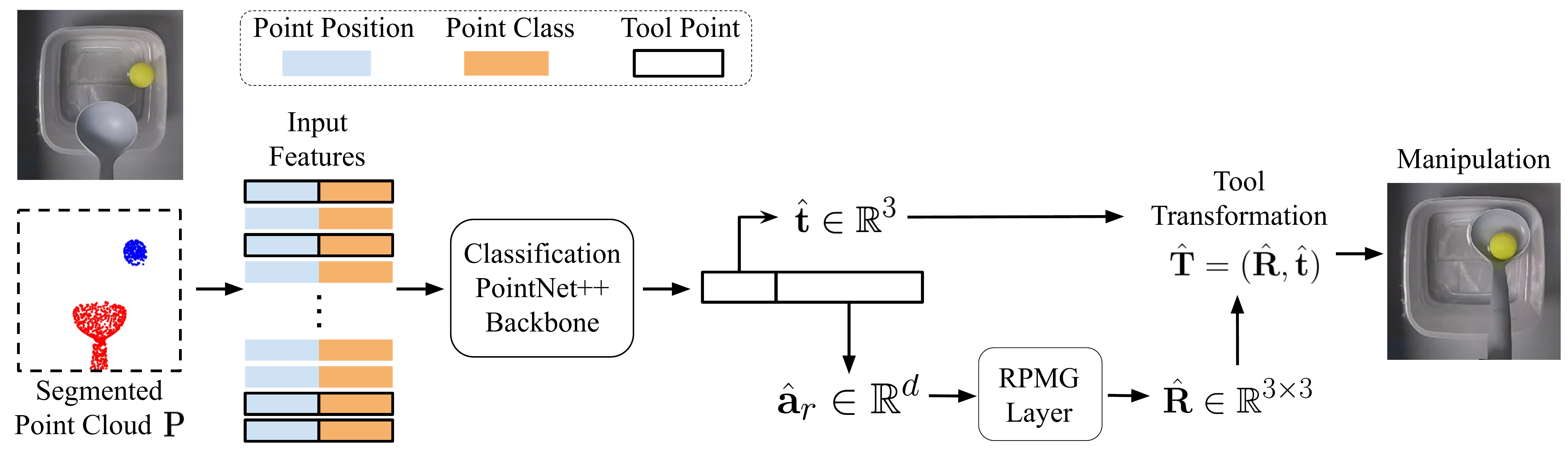}
  \caption{
  \hl{PCL Direct Vector baseline, adjusted to test different rotation representations. With a standard segmented point cloud as input, it uses a classification PointNet++ to output a single vector, split into a translation $\hat{\bt}$ and a rotation $\hat{\ba}_r$ component. For $\hat{\ba}_r \in \mathbb{R}^d$, we test 4D, 6D, 9D, and 10D rotation representations ($d \in \{4,6,9,10\}$), and use an RPMG layer to project $\hat{\ba}_r$ to a rotation matrix.}
  See Section~\ref{app:rpmg-rotations} \hl{for details.}
  }
  \vspace{0pt}
  \label{fig:rpmg}
\end{figure}

\hl{To implement this, we use a classification PointNet++ network which takes in the same segmented point cloud as input. Instead of the output as a vector in $\mathbb{R}^6$, as is the case for the PCL Direct Vector method, the output is a vector in $\mathbb{R}^{3 + d}$, split into the translation prediction $\hat{\bt} \in \mathbb{R}^3$ and a $d$-dimensional rotation vector $\hat{\ba}_r \in \mathbb{R}^d$. We then pass the $\hat{\ba}_r$ vector through the RPMG layer}~\cite{RPMG} \hl{to produce a (predicted) rotation matrix $\hat{\mathbf{R}} \in \mathbb{R}^{3\times 3}$. During backpropagation, the RPMG layer produces gradients through the rotation representation, by taking gradients on the $SO(3)$ manifold for the rotation representations. To reduce the chances of implementation errors, we directly reuse the layer from the RPMG}~\cite{RPMG} code.\footnote{\url{https://github.com/jychen18/RPMG}}
See Figure~\ref{fig:rpmg} \hl{for a visualization of the architecture.}

\hl{The RPMG layer introduces two hyperparameters, $\lambda$ and $\tau$. Following the RPMG paper, we fix $\lambda=0.01$ and adjust $\tau$ throughout training by initially setting it to $\tau_{\rm init} = 0.05$ and then increasing it to $\tau_{\rm converge} = 0.25$ at the end of 500 Behavioral Cloning training epochs.}

\hl{To train this model with the RPMG layer, we optimize the sum of translation and rotation losses. For translation, we use mean-square error, and for rotation, we follow the RPMG paper and minimize the Frobenius norm of the difference between the predicted and ground truth rotations: $\| \hat{\mathbf{R}} - \mathbf{R}^*\|_F$. We apply equal weight to the translation and rotation losses.}

\hl{We show the Behavioral Cloning results with different rotation representations in Table}~\ref{tab:rpmg-rotations}, \hl{and compare with the standard 3D axis-angle rotation representation and ToolFlowNet. The results suggest that none of the alternative rotation representations offer performance benefits. We have also tried using the point matching loss instead of adding separate MSE and Frobenius norm losses, but the results were worse and we do not report them.} 

\clearpage
\section{Physical Experiments}\label{app:physical}

In this section, we discuss our physical experiments in more detail and present new results with more general starting configurations.

\subsection{Physical Setup}

\begin{wrapfigure}{r}{0.40\textwidth}
\vspace{-12pt}
\centering
\includegraphics[width=0.40\textwidth]{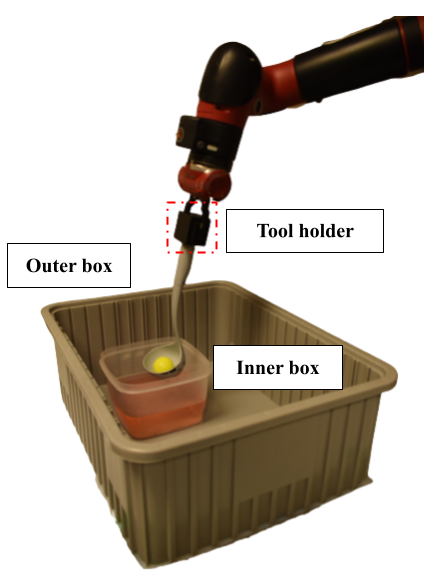}
\caption{
\small
Closeup of inner and outer box and the tool holder.
}
\vspace*{-10pt}
\label{fig:physical-closeup}
\end{wrapfigure}
The experimental setup consists of a Rethink Sawyer robot. We attach a standard consumer ladle to its gripper to make it feasible for the Sawyer to scoop a \hl{yellow} floating ping-pong ball.
We use Shining 3D EinScan-Pro to obtain the mesh of the ladle. We then convert this mesh to a point cloud that we query tool points from, while collecting ground-truth demonstrations and while running inference. A custom designed, 3D printed tool holder made of ABS plastic attaches the ladle to the end-effector.

An overhead Microsoft Kinect Azure camera \hl{continuously} queries depth and RGB images \hl{of the scene}, which we use to generate point clouds $\pcl$. \hl{Given the distinct yellow color of the target, we can segment the target points from the point cloud using HSV thresholding on the Kinect's RGB images. This gives us one of the two segmentation classes.} At each time step, ROS's \texttt{tf} functionality queries the transformation between the Sawyer's base frame and the end-effector link. We apply \hl{this transformation} to the scanned 3D model of the ladle to \hl{obtain a transformed model of the tool.} \hl{We sample points from this transformed tool model to obtain the tool points. Through this technique, we obtain the second segmentation class, pertaining to the tool points}. When collecting demonstrations, we track the changes in the pose of the ladle at consecutive time-steps to derive the tool \hl{flow}. These form the observation-action pairs to train \toolflow.

At the start of each demonstration and each test-time trial, we drop a yellow ping-pong ball in a translucent box in Figure~\ref{fig:physical-closeup} which contains water.

The water \hl{contains} red food coloring to provide better color contrast for accurate \hl{HSV} segmentation of the ping-pong ball. 
We tape the inner box within a larger box, which is the outer, gray box in Figure~\ref{fig:physical-closeup}; this helps to contain spills and to prevent the smaller box from sliding. The gray box we use is from MSC Industrial Direct Co. and is a 100 Lb Load Capacity Gray Polypropylene Dividable Container with dimensions 22.5 inches long, 17.5 inches wide, and 8 inches tall.

\subsection{Experiment Details}

\hl{The demonstrations only describe translation motions, and we use the Sawyer's impedance controller to avoid end-effector rotations.}
We will test the model's performance in the physical environment with rotations in future work. One author of this paper collected all the training demonstrations. 

\begin{figure}
    \centering
    \includegraphics[scale=0.30]{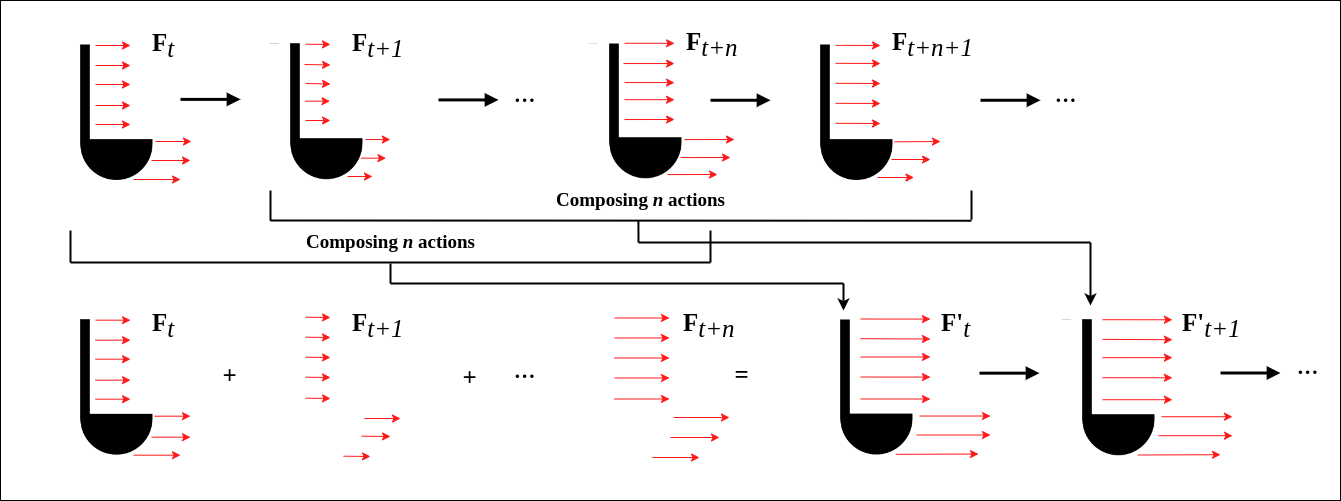}
    \caption{\hl{Visualization of the variable composing technique used in the (translation-only) physical experiments. We compose the ground-truth action targets until the composed flow vectors are at least} \SI{1}{\centi\meter}. \hl{In the figure, for time-step $t$, $n$ actions are composed together to generate the variably composed flow represented as $\flow'_{\rm t}$, which replaces the flow $\flow_{\rm t}$, which has a magnitude less than} \SI{1}{\centi\meter}. \hl{Similarly, for time step $t+1$, $n$ actions are composed together to generate the new flow, $\flow'_{\rm t+1}$, replacing the original flow, $\flow_{\rm t+1}$.}
    }
    \label{fig:physical-compose-actions}
\end{figure}

During initial physical tests with collecting demonstration data, we noticed that ground-truth translations were roughly \SIrange{2}{3}{\milli\meter} in magnitude, which could result in small and jerky robot motions at test time from a learned policy. Thus, we \hl{compose the ground truth actions until their magnitude is at least} \SI{1}{\centi\meter}. \hl{These composed actions then become the ground truth training targets for the point cloud observations at each time step.} Figure~\ref{fig:physical-compose-actions} \hl{depicts the variable composing method. In this example for the flow at time step $t$ and $t+1$, we compose $n$ actions to generate the flow $\flow'_{\rm t}$ and $\flow'_{\rm t+1}$ respectively, which both have a magnitude of at least} \SI{1}{\centi\meter}.  \hl{While training ToolFlowNet on the physical experiment data, we scale the ground truth actions so that their values roughly lie in the range of -1 to +1, similar to the protocol followed for the simulation experiments} (see Section~\ref{app:scaling-targets}). \hl{At test time, we downscale the actions predicted by the network with the same factor}.

The Sawyer is controlled by a computationally lightweight computer, which lacks the ability to run GPU intensive inference using the trained \toolflow model. Furthermore, the Sawyer is controlled using ROS 1, which runs on Python 2, whereas we train \toolflow using Python 3. At each time step, we therefore send the point cloud observations to a separate, more powerful GPU-enabled machine with Python 3 to run inference using \toolflow and generate the necessary action commands.
To interface the control computer with the GPU-enabled machine, we utilize Python bindings from ZeroMQ~\cite{zeromq}, called \texttt{pyzmq} to create a SSH tunnel between the two machines. 

\subsubsection{Experiment Protocol}

\begin{figure}[t]
  \centering
  \includegraphics[width=1.00\textwidth]{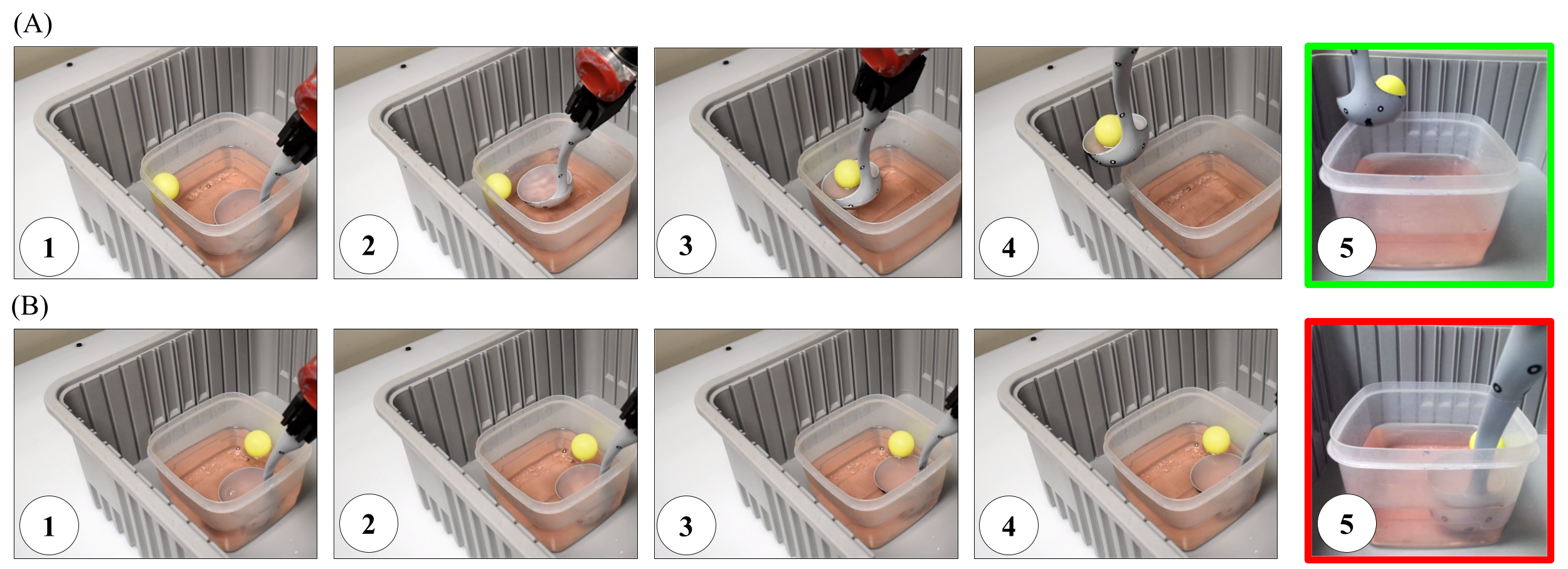}
  \caption{Subsampled frames from the testing trials during the physical experiments. Frame 1 shows the starting location of the target. The alternative camera view in frame 5 shows the height to which the robot lifts the target at the end of the trial. A) Frames from a scooping success, where the model successfully locates the ball (frame 2) and then manages to raise it above the top of the inner, translucent box (frame 4). Frame 5 shows the side view, where it is apparent that the robot has lifted the target, well over the top of the inner box. B) Frames from a scooping failure, where the robot was not able to locate or lift the ball to the top of the inner box \hl{due to collisions with the bottom right corner of the inner box}. Frame 5 \hl{in the bottom row} shows the ladle still submerged in the water.}
  \vspace{-0pt}
  \label{fig:physical-success-failure}
\end{figure}

We judge the performance of \toolflow on whether the Sawyer successfully scoops the ping-pong ball (\ie the target) out of the water \hl{without colliding with the rest of the experimental setup}.

In Section~\ref{sec:experiments}, we report the success rate of \toolflow in experiments where the target was dropped at some arbitrary location inside the smaller inner box. At the start of each trial, we initialize the ladle such that its bowl is just under the surface of the water. \hl{The robot then executes actions predicted by \toolflow, in order to scoop the ball out of the water. The robot scoops the ball in an average of 17 time steps.}

In Section~\ref{sec:experiments}, all the failures occur when the Sawyer repeatedly pushes its ladle against the walls of the inner box. Subsampled frames from a successful trial and a \hl{collision} failure are shown in Figure~\ref{fig:physical-success-failure}, rows (A) and (B), respectively. Collision failures are better conveyed through videos and can be found on the project website: \url{https://tinyurl.com/toolflownet}.

\end{document}